\documentclass{article}
\usepackage[T1]{fontenc}
\usepackage{microtype}
\usepackage{graphicx}
\usepackage{subfigure}
\usepackage{booktabs} %
\usepackage{multirow}
\usepackage{soul}
\usepackage[hyphens]{url}
\usepackage{wrapfig}

\usepackage{hyperref}

\usepackage[accepted]{icml2024}

\usepackage{amsmath}
\usepackage{amssymb}
\usepackage{mathtools}
\usepackage{amsthm}
\usepackage{xspace}
\usepackage{listings}
\usepackage{caption}
\usepackage[capitalize,noabbrev]{cleveref}

\usepackage{titlesec}

\theoremstyle{plain}

\theoremstyle{definition}

\theoremstyle{remark}

\newcommand{\method}{Contextualized Critiques with Constrained Preference Optimization\xspace}
\newcommand{\methodshort}{C3PO\xspace}
\NewDocumentCommand\emojibooks{}{
    \includegraphics[scale=1]{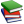}
}

\usepackage[textsize=tiny]{todonotes}

\icmltitlerunning{RLVF: Learning from Verbal Feedback without Overgeneralization}

\begin{document}

\twocolumn[
\icmltitle{RLVF: Learning from Verbal Feedback without Overgeneralization}

\icmlsetsymbol{equal}{*}

\begin{icmlauthorlist}
\icmlauthor{Moritz Stephan}{yyy}
\icmlauthor{Alexander Khazatsky}{yyy}
\icmlauthor{Eric Mitchell}{yyy}
\icmlauthor{Annie S Chen}{yyy}
\icmlauthor{Sheryl Hsu}{yyy}
\icmlauthor{Archit Sharma}{yyy}
\icmlauthor{Chelsea Finn}{yyy}
\end{icmlauthorlist}

\icmlaffiliation{yyy}{Department of Computer Science, Stanford University, CA, USA}

\icmlcorrespondingauthor{Moritz Stephan}{moritzst@stanford.edu}

\icmlkeywords{Machine Learning, ICML}

\vskip 0.3in
]

\printAffiliationsAndNotice{\hspace{-4mm}} %

\begin{abstract}
The diversity of contexts in which large language models (LLMs) are deployed requires the ability to modify or customize default model behaviors to incorporate nuanced requirements and preferences. A convenient interface to specify such model adjustments is high-level verbal feedback, such as “Don’t use emojis when drafting emails to my boss.” However, while writing high-level feedback is far simpler than collecting annotations for reinforcement learning from human feedback (RLHF), we find that simply prompting a model with such feedback leads to \textbf{overgeneralization} of the feedback to contexts where it is not relevant. We study the problem of incorporating verbal feedback without such overgeneralization, inspiring a new method {\method} (\methodshort). {\methodshort} uses a piece of high-level feedback to generate a small synthetic preference dataset specifying how the feedback should (and should not) be applied. It then fine-tunes the model in accordance with the synthetic preference data while minimizing the divergence from the original model for prompts where the feedback does not apply. Our experimental results indicate that our approach effectively applies verbal feedback to relevant scenarios while preserving existing behaviors for other contexts. For both human- and GPT-4-generated high-level feedback, \methodshort effectively adheres to the given feedback comparably to in-context baselines while reducing overgeneralization by 30\%.

\end{abstract}

\section{Introduction}
\label{Intro}

\begin{figure*}
    \centering
    \includegraphics[width=\textwidth]{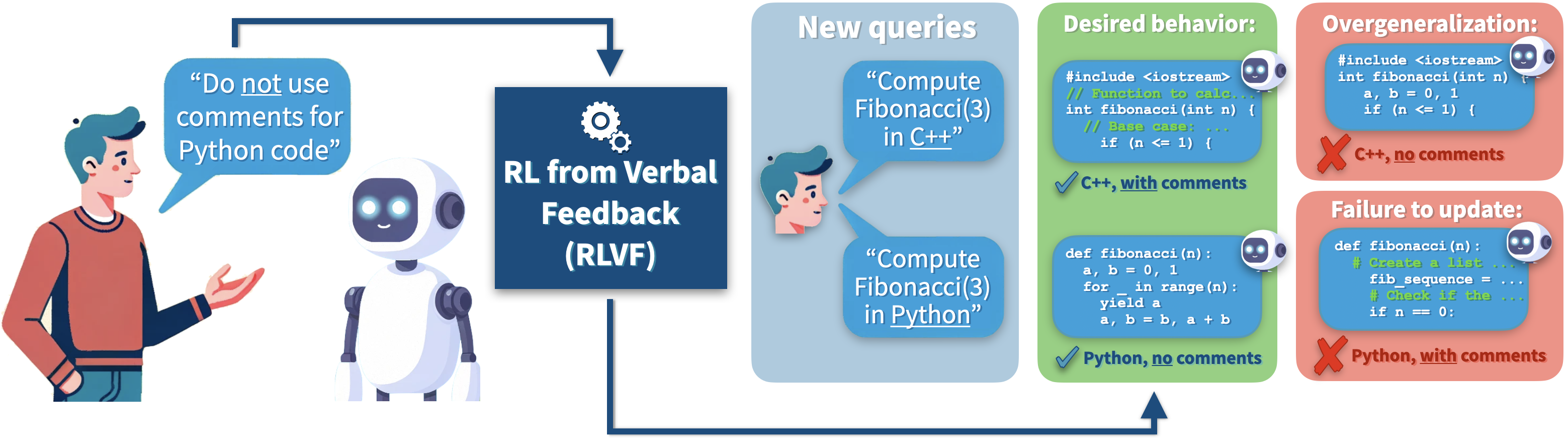}
    \caption{We consider the problem of leveraging \textbf{high-level, verbal} feedback (left) to refine model behaviors (center). Prior approaches often struggle to appropriately update the model, leading to either failure to adhere to the feedback or overgeneralization (right).}
    \label{fig:teaser}
\end{figure*}

With the increasingly widespread adoption of large language models (LLMs) across diverse industries and individuals, the ability to align them with high-level human feedback for a specific user or use-case becomes increasingly important. While LLM users often want the model to adhere to broad principles at all times, such as producing fluent text, individual users and use-cases have more nuanced preferences. For example, a user may request the LLM to write more concise work emails but more detailed personal emails, making feedback context dependent. Tailoring models to accommodate such preferences is challenging: it requires extensive resources to gather preferences in all different contexts and fine-tuning the model in one context can unpredictably impact model behavior in other contexts. We study the problem of adapting models using verbal feedback that is fast and easy for people to provide (see Fig.~\ref{fig:teaser}).

Common approaches to incorporating feedback, such as supervised context distillation (SCD) or reinforcement learning from human feedback (RLHF), use example-level supervision via either supervised completions or preference labels. Such methods require a corpus of user-provided (preference-)data, which can be costly and cumbersome to obtain. 
Additionally, they do not constrain model behavior outside the context that the feedback may apply, so the LLM might adapt its behavior in unintended ways, e.g. output a more verbose work email when the preference only applies to personal emails. 
Verbal feedback is far easier and faster for humans to provide. To this end, another common approach is to incorporate such verbal feedback into the prompt, potentially through an iterative process to continually add additional points of feedback. However, this approach requires re-using the prompt in all future queries. As more pieces of feedback accumulate, long prompts containing many context-dependent feedbacks can make inference expensive; further, identifying which pieces of feedback should apply in a given context can become difficult.

We aim to adapt LLMs such that, when provided with a single sentence that specifies feedback, the model can discern which situations the feedback applies to and appropriately incorporate the feedback in future outputs. We propose \method (\methodshort), where we first synthetically generate hypothetical prompts in-scope and out-of-scope for the feedback. We then sample original completions to these prompts, without the feedback applied, as well as revised completions in line with the feedback. Importantly, we utilize the strong priors of existing instruction-tuned LLMs in this process and therefore do not require any additional human supervision. We then introduce a new objective to fine-tune the LLM's response behavior. One naive approach might use original and revised completions for prompts to maximize the implicit reward of a preference model (e.g. using direct preference optimization~\citep{rafailov2023direct}). However, this objective does not capture the need to leave model behavior unchanged for non-relevant prompts. Instead, \methodshort jointly maximizes the implicit reward for in-scope prompts and minimizes standard cross-entropy loss between the logits of the base and fine-tuned model for out-of-scope prompts. Including the latter loss in the objective adjusts the LLM's responses to prompts where the feedback is relevant, while preserving its behavior in contexts where the feedback should not be applied.

Our main contribution is \methodshort, a new method for learning from verbal feedback that selectively adapts the LLM's behavior based on the context of the feedback. This novel synthetic data generation scheme and fine-tuning objective enables an LLM to extrapolate single-sentence feedback to new situations.   
Across numerous examples of feedback generated by humans and GPT-4, we find that \methodshort accurately applies the feedback to relevant prompts and importantly, substantially reduces unintended behavior changes in scenarios where the feedback is not applicable, outperforming prior methods by over 10\% when both criteria are considered.
By providing adaptability to verbal feedback while reducing overgeneralization of such feedback, our work may help enhance the utility of LLMs in diverse use-cases.

\section{Related Work}

Improving language or dialogue systems from feedback has been studied in the context of various types of feedback, including learned or \citep{walker2000reinforcement,bohm-etal-2019-better} or heuristic \citep{li-etal-2016-deep} rewards on individual model outputs, preferences or rankings over pairs or sets of model samples \citep{ziegler2020finetuning,ouyang2022training,rafailov2023direct}, and natural language feedback on model outputs or behaviors \citep{li2017dialogue}. Natural language feedback or corrections on individual model outputs have been used to improve performance in code generation \citep{austin2021program,chen2023improving}, dialogue \citep{li2017dialogue,hancock-etal-2019-learning,shi2022life}, and summarization \citep{scheurer2023training}. Feedback or critiques are typically used to refine model outputs during generation, iterating on or refining the model's initial response before outputting a final answer. Recent work has emphasized \textit{self-}refinement, where an LLM generates its own feedback \citep{madaan2023selfrefine,large_huang_2023,automatically_pan_2023}. Some studies have shown that the final outputs from such (self-)refinement methods can be distilled back into the model, improving its base performance without requiring iterative refinement during sampling at test time \citep{principledriven_sun_2023,self_lu_2023,yu2023teaching,yuan2024selfrewarding,yang2024rlcd}.

Most relevant to the present work are studies leveraging natural language feedback to refine general model behaviors, rather than iteratively improving a single model output. Constitutional AI \citep{constitutional_bai_2022} uses an LLM to generate synthetic training data that encourages an LLM to follow high-level rules written by a human; \citet{glaese2022improving} uses a similar approach to instill various rules into a pre-trained LLM. Context distillation \citep{askell2021general,learning_snell_2022} is another approach to controllability that distills the behavior of the LLM when conditioned on a piece of feedback back into the LLM \textit{without} the feedback present, essentially `baking in' the feedback. However, these approaches to controllability have mostly been used to instill universal behavioral changes (i.e., rules that should always be adhered to). Relatively fewer works have studied conditional or context-dependent rule following (though \citet{clark2021transformers} study adherence to synthetic rules for logical reasoning and commonsense). In concurrent work, \citet{castricato2024suppressing} also utilize model completions and revisions to generate synthetic preference pairs; they use this technique to train a language model to better follow instructions that specifically request \textit{avoiding} a given topic.

A related problem is \textit{model editing} \citep{Sinitsin2020Editable,editing_decao_2021,fast_mitchell_2021,locating_meng_2022}, which studies interventions to pre-trained models that should only apply in a relatively small neighborhood around the `model edit' (desired intervention). Most work in model editing studies corrections to factual or reasoning errors. However, \citet{memorybased_mitchell_2022} study edits that adjust the sentiment of a dialogue model for a single topic, and \citet{fixing_murty_2022} show edits (or `patches') to sentiment classification or relation extraction models. \citet{editing_mao_2023} extend this work by editing model behavior for a single topic according to three categories of personality traits. \citet{akyurek-etal-2023-dune} and \citet{hewitt2024model} study model edits aimed at debiasing LLMs; \citet{hewitt2024model} also study factual edits and corrections to syntactic errors. In contrast, our work performs general behavioral edits to pre-trained LLMs, rather than edits of a specific type or for a specific context.

\section{Preliminaries}

We first outline two approaches for updating LLMs with high-level verbal feedback: supervised context distillation and preference-based reinforcement learning (PbRL).

\textbf{Supervised context distillation.} A simple but effective method to updating language models from feedback is supervised context distillation (SCD; \citet{askell2021general}), which incorporates a textual context $z$ containing a general principle (e.g., ``Always be nice!'') or information (e.g., ``Assume the US president is Joe Biden.'') into a model's behavior. SCD `distills' the behavior that a human or an LLM $\pi_0$ would produce when conditioned on both a user query $x$ and the context $z$ into the LLM without the context present. That is, from a dataset of unlabeled user queries or prompts $\mathcal{D}_u = \{x_i\}$, a distillation target $y_i$ is either written by a human or generated by the LLM $\pi_0$ for each $x_i$ as $y_i \sim \pi_0(\cdot \mid x, z)$. The language model $\pi_{\theta}$ is produced from supervised fine-tuning with the negative log likelihood loss, using the synthetic supervision targets:
\begin{equation}
    \mathcal{L}_\text{SFT}(\mathcal{D}) = -\mathbb{E}_{x,y\sim \mathcal{D}}\log \pi_\theta(y \mid x),
\end{equation}
where $\pi_\theta$ is typically initialized as $\pi_0$.

\textbf{Preference-based reinforcement learning.} Preference-based reinforcement learning (PbRL; \citep{BusaFekete2014,ruiz2023dueling}) is the most widely-used approach to updating language models from feedback by leveraging preferences over pairs\footnote{Rankings over larger sets of responses can also be used, but we use pairs for simplicity.} of LLM-generated responses $y,y'$ to an input $x$. The responses $y,y'$ are typically sampled from a language model $\pi_0$ fine-tuned with SCD or a similar objective \citep{ziegler2020finetuning,constitutional_bai_2022,ouyang2022training}. The input $x$ may be an instruction, document to summarize, or dialogue history, for example. Given an input $x$, responses $y,y'$, %
an annotator (either a human or an LLM) labels which response is better, %
ultimately producing a dataset $\mathcal{D}_\text{pref} = \{x_i, y_i^+, y_i^-\}$, where $y_i^+$ is preferred to $y_i^-$ for query $x_i$, as judged by the annotator. %

\begin{figure}
    \centering
    \includegraphics[width=\columnwidth]{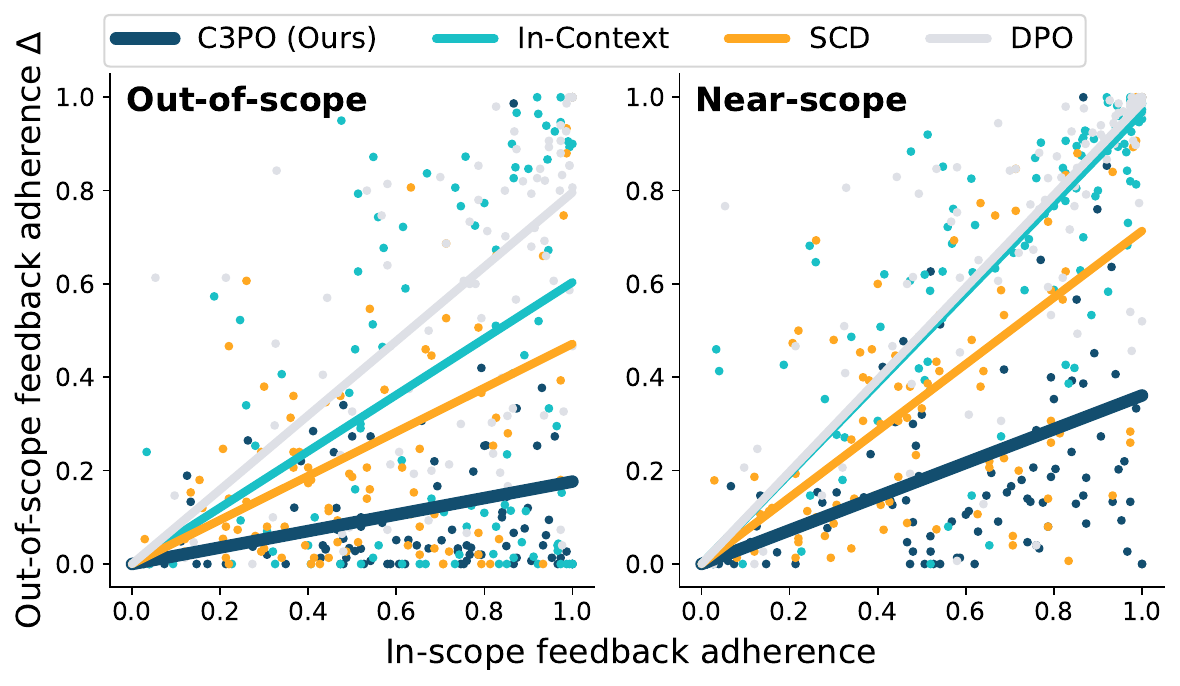}
    \caption{{\methodshort} mitigates the \textbf{overgeneralization problem} when learning from high-level feedback. For existing approaches for incorporating high-level feedback, high feedback adherence on in-scope prompts (x axis) strongly predicts a large change in behavior for out-of-scope prompts (y axis), which is undesirable. In contrast, our approach {\methodshort} decreases the rate at which out-of-scope behavior is affected as in-scope feedback adherence improves. Lines of best fit are computed with linear orthogonal regression.
    }
    \label{fig:overgeneralization}
\end{figure}

The dataset $\mathcal{D}_\text{pref}$ is used to learn a parameterized \textit{reward model} $r_\phi: \mathcal{X} \times \mathcal{Y} \rightarrow \mathbb{R}$ that assigns scalar goodness scores to individual input-response pairs. The most common objective for training a reward model is maximum likelihood in the Bradley-Terry choice model \citep{bradley1952rankanalysis}:
\begin{align}
    \mathcal{L}_\text{BT}(\phi) &= -\mathbb{E}_{x,y^+,y^-}\log p_{r_\phi}(y^+ \succ y^- \mid x,y^+,y^-) \\
    &= -\mathbb{E}_{x,y^+,y^-}\log \sigma\left(r_\phi(x,y^+) - r_\phi(x,y^-)\right)
\end{align}

Early methods for fine-tuning LLMs from human preferences followed the reward modeling stage with a policy optimization stage aimed at finding a language model policy $\pi_\theta$ that produces high-reward responses without deviating excessively from the LLM that generated the responses in $\mathcal{D}_\text{pref}$~\citep{schulman2017proximal,ziegler2020finetuning}. More recently, direct preference optimization (DPO; \citet{rafailov2023direct}) shows that the optimal policy can be extracted from the learned reward in closed form, avoiding the need for iterative policy search. Due to its simplicity and computational efficiency, we use the DPO algorithm for learning from preference data in this work. DPO directly optimizes the language model policy from preferences using the loss:
\begin{equation}
    \resizebox{\columnwidth}{!}{$\mathcal{L}_\text{DPO}(\mathcal{D}) = -\mathbb{E}_{x,y^+,y^-\sim\mathcal{D}}\log \sigma \left(\log\frac{\pi_\theta(y^+ \mid x)}{\pi_0(y^+ \mid x)} - \log\frac{\pi_\theta(y^- \mid x)}{\pi_0(y^- \mid x)}\right)$.}
\end{equation}
Our approach leverages PbRL to update a language model from high-level verbal feedback and does not assume that a preference dataset is directly available at the outset.

\begin{table}
    \small
    \centering
    \begin{tabular}{p{1.7cm}p{5.5cm}}
        \toprule
        \textbf{Feedback} & For specific Python coding questions, respond with only a code snippet and no explanations before or after the snippet. \\
        \midrule
        \midrule
        Categories & Data structures in various languages; Statistical computing in different environments \\
        \midrule
        \multirow{2}{*}{\shortstack[l]{In-scope\\prompts}} & Write a basic Queue class in Python. \\
        & How can I implement backprop in Python? \\
        \midrule
        \multirow{2}{*}{\shortstack[l]{Out-of-scope\\prompts}} & When did the Haitian revolution begin? \\
        & Can you explain relativity in a paragraph? \\
        \midrule
        \multirow{2}{*}{\shortstack[l]{Near-scope\\ prompts}} & What good C++ libraries are there for trees? \\
        & Is Python or Julia more popular for NLP? \\
        \bottomrule
    \end{tabular}
    \caption{An example feedback, prompt categories, and in-scope, out-of-scope, and near-scope prompts for each category.}
    \label{tab:example-feedback}
\end{table}

\section{Reinforcement Learning from Verbal Feedback using \method}
\label{sec:method}
Our goal is to enable adaptation to high-level verbal feedback without extensive human annotation. The verbal feedback $z$ corresponds to short passages of natural language text that describe the user's feedback on the model's current behavior. Unfortunately, na\"ively applying existing approaches to this problem leads the model to \emph{overgeneralize}, applying the feedback both when it should be applied and when it should not. 
Our aim is to develop a method capable of only applying feedback where it is appropriate. Starting with a base language model $\pi_{0}$, our approach, \method, uses a strong general-purpose model (such as GPT-4) to translate a piece of verbal feedback $z$ into a dataset which can be finetuned on to address feedback adherence without overgeneralization. This dataset will be an aggregation of three sub-datasets, which all serve a unique purpose. The first component of this dataset, $\mathcal{D}_\text{in-scope}$, exists to demonstrate the desired change of behavior. Next, we have $\mathcal{D}_\text{out-of-scope}$, which allows us to maintain our behavior outside the scope of the feedback. Lastly, we have $\mathcal{D}_\text{near-scope}$, which is adversarially designed to refine our model's understanding of where it is appropriate to apply the feedback. To update the model, we jointly train on $\mathcal{D}_\text{in-scope}$ with PbRL and the unison of $\mathcal{D}_\text{out-of-scope}$ and $\mathcal{D}_\text{near-scope}$ with simple SFT. We now describe our dataset generation procedure in more detail.

\begin{figure}
    \centering
    \includegraphics[width=\columnwidth]{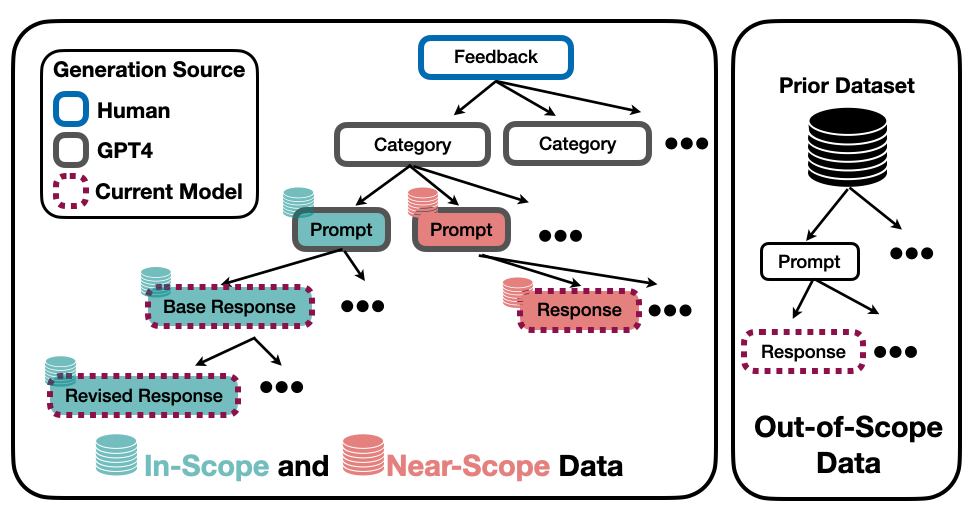}
    \caption{\textbf{\methodshort Data Generation Scheme.} Given human feedback, \methodshort begins by generating a set of categories of prompts where the feedback may be relevant using GPT-4. GPT-4 then generates in-scope prompts $x_i^\text{in-scope}$ and near-scope prompts $x_i^\text{near-scope}$. A set of out-of-scope prompts $x_i^\text{out-of-scope}$ is also taken from a prior dataset. The current model then generates a baseline response to each of these, giving $y_i^-$, $y_i^{\text{near-scope}}$, $y_i^{\text{out-of-scope}}$, respectively. We also prompt the current model to revise $y_i^-$ to incorporate the feedback, giving a revised response $y_i^+$.
    This data generation scheme is the first stage of \methodshort--autonomously generating fine-tuning datasets $\mathcal{D}_\text{in-scope}$, $\mathcal{D}_\text{near-scope}$ and $\mathcal{D}_\text{out-of-scope}$, the latter two of which are used to prevent overgeneralization on irrelevant tasks.}
    \label{fig:data-generation}
\end{figure}

\textbf{Translating high-level verbal feedback into a fine-tuning dataset.} PbRL algorithms like DPO optimize a model's behavior only for a given distribution of inputs; thus, to incorporate a piece of feedback $z$, we must first determine the distribution of model inputs where the model's behavior should change. Given a piece of feedback $z$, {\methodshort} uses GPT-4 to first generate a set of $K$ \textit{categories} of prompts where the feedback could apply. GPT-4 then generates $M$ prompts $x_i^\text{in-scope}$ ($\frac{M}{K}$ for each category) where the feedback applies. However, beyond accommodating feedback for in-scope prompts, we must also avoid overgeneralization of the feedback to prompts where it does not apply. We therefore generate a set of $M$ prompts $x_i^\text{near-scope}$ in each category that are superficially related to the feedback in some way (lexically, semantically), but are not actually inputs where the model's behavior should change. Finally, we use a fixed set of $M$ feedback-independent prompts $x_i^\text{out-of-scope}$ to avoid degradation of completely unrelated model behaviors.\footnote{We sample these prompts randomly from the Open Instruction Generalist dataset \citep{oig}.} See \cref{tab:example-feedback} for an example feedback, categories, and prompts and \cref{fig:data-generation} for a summary on the data generation scheme.

\begin{figure*}
    \centering
    \includegraphics[width=0.9\textwidth]{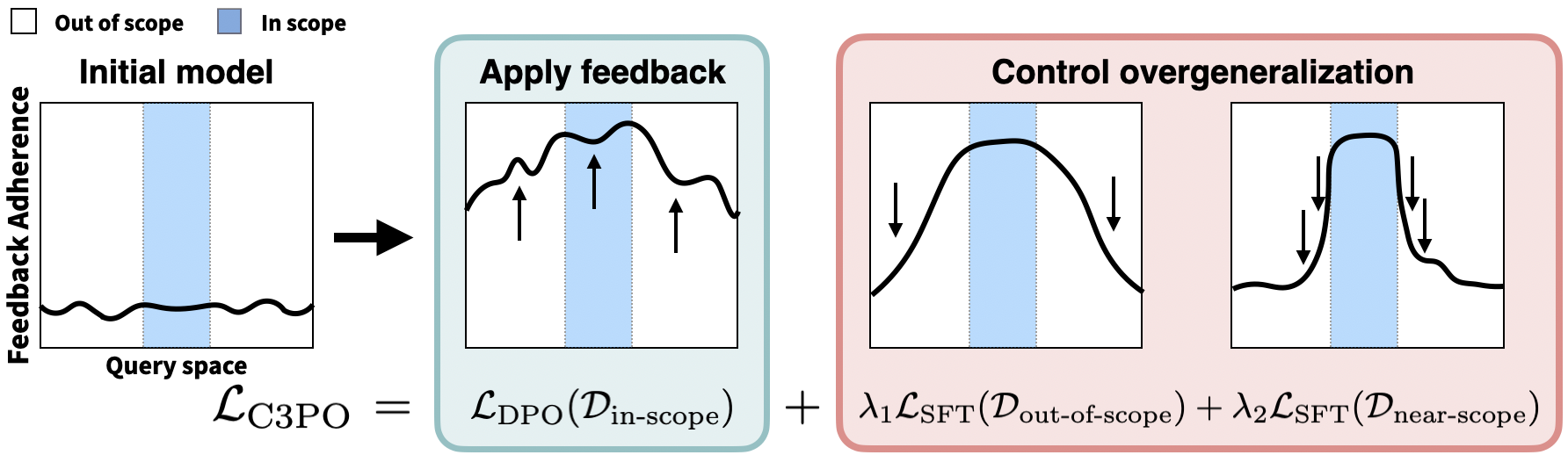}
    \caption{\textbf{{\methodshort} Fine-Tuning Objective.} \methodshort facilitates feedback adherence for relevant prompts by fine-tuning with DPO on the generated in-scope data while minimizing overgeneralization through SFT losses on the generated out-of-scope and near-scope data, which regularizes model behavior towards the original model for feedback-irrelevant prompts.}
    \label{fig:objective}
\end{figure*}

To capture the desired delta in behavior denoted by the feedback, we generate the dataset of preference pairs using the in-scope prompts $\mathcal{D}_\text{in-scope} = \{x_i^\text{in-scope}, y_i^+, y_i^-\}$. $y_i^-$ is generated by the language model that originally received the feedback, i.e., we have $y_i^-\sim \pi_0(\cdot \mid x_i^\text{in-scope})$. To generate $y_i^+$, the language model is then prompted to revise $y_i^-$ to incorporate the feedback, i.e., we have $y_i^+\sim \pi_0(\cdot \mid x_i^\text{in-scope},y_i^-,z)$. See \cref{sec:appendix-prompts} for the complete prompt format. Thus to the extent that the model $\pi_0$ can correctly interpret the given feedback, the generated preference data represents the desired `delta' in behavior described by the feedback $z$. To control for model degradation on out-of-scope prompts, we populate $\mathcal{D}_\text{near-scope}=\{x_i^\text{near-scope}, y_i^\text{near-scope}\}$ and $\mathcal{D}_\text{out-of-scope}=\{x_i^\text{out-of-scope}, y_i^\text{out-of-scope}\}$ with the respective prompts and corresponding completions sampled from the initial language model $\pi_0$. These datasets encode the behaviors that we want to \textit{preserve} after incorporating $z$.

\textbf{Fine-tuning using the synthetic data.}
Using our synthetically-generated datasets, we now fine-tune the model $\pi_0$ using a combined loss, as shown in \cref{fig:objective}, that both incorporates feedback on relevant prompts and discourages model degradation on feedback-irrelevant prompts:
\begin{multline}
    \mathcal{L}_\text{\methodshort} = \overbrace{\mathcal{L}_\text{DPO}(\mathcal{D}_\text{in-scope})}^\text{Apply feedback} \\ + \underbrace{\lambda_1 \mathcal{L}_\text{SFT}(\mathcal{D}_\text{out-of-scope}) + \lambda_2 \mathcal{L}_\text{SFT}
    (\mathcal{D}_\text{near-scope})}_\text{Control model degradation}.
\end{multline}

\textbf{Interpreting the {\methodshort} loss.} While the $\mathcal{L}_\text{SFT}$ losses simply regularize the updated model $\pi_\theta$ toward the original model $\pi_0$ for prompts not relevant to the feedback, the result of learning from the {\methodshort} synthetic preference dataset for in-scope inputs is less obvious. {\methodshort} generates what we refer to as \textit{synthetic two-policy preference data} $(x,y^+,y^-)$, where $y^+$ is always preferred to $y^-$. These preference tuples are constructed by simply sampling $y^-$ from a policy $\pi^-$ (the baseline model $\pi_0$) and $y^+$ from a different policy $\pi^+$ (the baseline model $\pi_0$ prompted to revise a baseline model response using the feedback).\footnote{Some existing work \citep{yang2024rlcd,intel2023neuralchat} shows that synthetic two-policy preference data can produce useful policies.} Unlike preference datasets scored by a black-box human or AI annotator, we can express the optimal policy learned from such preference data in terms of the data-generating policies.

We show in \cref{sec:math} that such synthetic two-policy preference data satisfies the Bradley-Terry (BT) preference model \citep{bradley1952rankanalysis}, which assumes that preference data $(x,y,y')$ are constructed according to some unknown scoring function $r^*$ as $p(y \succ y' | x) = \sigma \left(r^*(x,y) - r^*(x,y')\right)$. We show that two-policy preference data adheres to the BT model with
\begin{equation}
    r^*_\text{2p}(x,y) = \log \frac{\pi^+(y | x)}{\pi^-(y | x)}.
\end{equation}
Further, we show that when using $\pi^-$ as the reference model for PbRL, the optimal policy corresponds to
\begin{equation}
    \label{eq:2p-policy}
    \pi^*_\text{2p}(y | x) \propto \left(\frac{\pi^+(y | x)\phantom{^{1-\beta}}}{\pi^-(y | x)^{1-\beta}}\right)^\frac{1}{\beta}.
\end{equation}
Notable special cases of $\pi^*_\text{2p}$ are the geometric mean of $\pi^+$ and $\pi^-$ for $\beta=2$ and simply $\pi^+$ for $\beta=1$. For $\beta<1$, we interpret $\pi^*_\text{2p}$ as returning a temperature-sharpened version of $\pi^+$, but with a penalty on responses assigned high probability under $\pi^-$ (i.e., responses that respond to the user but fail to adhere to the feedback). See \cref{sec:math} for visualization of $\pi_\text{2p}^*$ with various $\beta$ in a synthetic setting.

\section{Experiments}

Our experiments are intended to answer several research questions about learning from verbal feedback. We first investigate the question: to what extent does the overgeneralization problem occur for existing methods for learning from verbal feedback, and does {\methodshort} mitigate this effect? Next, we study whether simple modifications to standard approaches to incorporating feedback with prompting or supervised context distillation effectively mitigate overgeneralization. Further, we study whether or not {\methodshort} can learn multiple pieces of feedback as well as the impact of the specific choice of the form of the constraint loss used in {\methodshort}. Before discussing the results of these experiments, we elaborate the datasets, evaluation metrics, and baseline methods used in our experiments.

\textbf{Datasets.} Our feedback dataset is composed of 100 pieces of feedback, where half are written by the authors\footnote{One piece of feedback was inspired by a suggestion from social media.} and half are generated by GPT-4 using the prompt provided in \cref{sec:appendix-prompts}. All pieces of feedback are designed to apply only in some contexts; see \cref{tab:example-feedback} for examples. For {\methodshort} and the \texttt{SCD + Negatives} baseline, the datasets $\mathcal{D}_\text{near-scope}$ and $\mathcal{D}_\text{out-of-scope}$, each containing out-of-scope prompts and corresponding baseline model completions used for regularization, are sampled according to the {\methodshort} procedure in \cref{sec:method}. For each piece of feedback, $|\mathcal{D}_\text{near-scope}|=|\mathcal{D}_\text{out-of-scope}|=960$. We sample the prompts for $\mathcal{D}_\text{out-of-scope}$ from the Open Instruction Generalist (OIG) Dataset \citep{oig} which contains a mix of diverse prompts ranging from math to QA and chat. Within each of the 32 categories of prompts, we randomly select $80\%$ to be used for training and validation and the remainder are used for testing.
\begin{figure}
    \centering
    \includegraphics[width=\columnwidth]{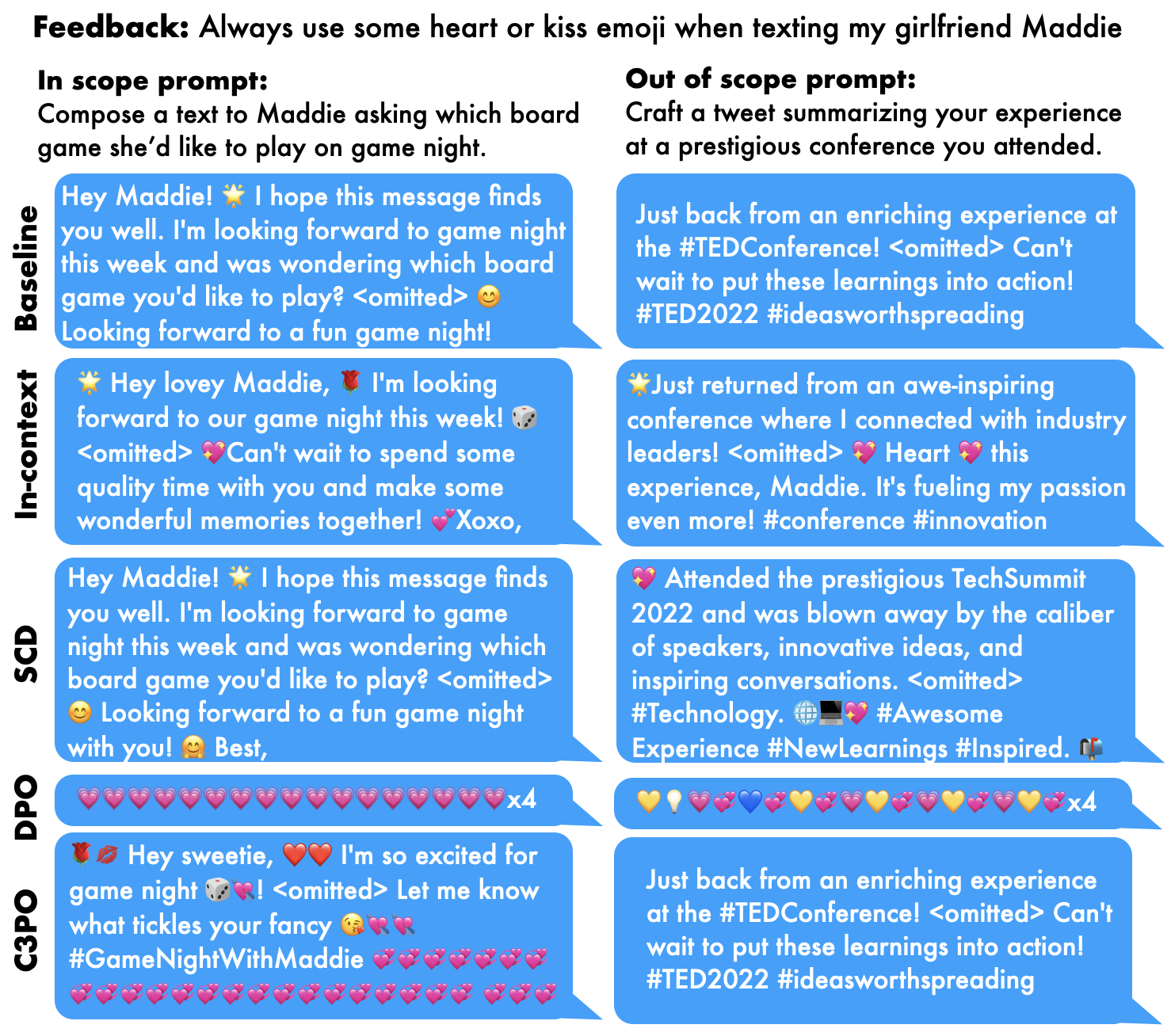}
    \caption{Sample responses from {\methodshort} and each baseline for an in-scope and out-of-scope prompt. Only {\methodshort} correctly adheres to the feedback for the in-scope input and ignores the feedback for the out-of-scope input.}
    \label{fig:feedback_samples}
\end{figure}

\textbf{Evaluation metrics.} Our evaluations are constructed to compare the feedback adherence of the baseline model with the model after learning from the feedback; we evaluate this change in behavior for both in-scope and out-of-scope prompts. For in-scope prompts, our goal is to increase feedback adherence, while for out-of-scope prompts, our goal is to preserve the rate of feedback adherence of the original model (that is, leave the baseline model unchanged). We measure feedback adherence in two ways, heuristically and with GPT-4. For 14 of the human-generated pieces of feedback such as modifying response length or the inclusion of certain words, manually crafted heuristic rules are sufficient to reliably measure which of two responses better adheres to a given piece of feedback. For a prompt $x$, a model output $y$, and a baseline response $\bar y$ from the baseline model, the heuristic scoring function produces a \textbf{feedback score} $h(x,y,\bar y)$. This scoring function intuitively scores whether the feedback adherence of the adapted model response is better than, equal to, or worse than the baseline model response. $h(x,y,\bar y)$ takes a value of 1 if $y$ incorporates the feedback and $\bar y$ does not, a value of 0 if both responses adhere to the feedback or neither response adheres to the feedback, and a value of -1 if the baseline response $\bar y$ adheres to the feedback and the adapted model response $y$ does not. In contrast, most pieces of feedback, such as requesting the usage of more metaphors or less aggressive speech, require qualitative evaluation. In these cases, we measure relative feedback adherence using GPT-4. For a prompt $x$, a model output $y$, and a baseline response $\bar y$ from the baseline model, we prompt GPT-4 to output a preference score when comparing two responses using the prompt in \cref{sec:appendix-prompts}, producing a feedback score $g(x,y,\bar y)$ scaled to be in the range $[-1, 1]$. This score measures the extent to which a given response adheres to the feedback better than the baseline response; a score of $1$ denotes that the trained model response adheres to the feedback much better than the baseline model and $-1$ denotes the reverse. We use these metrics (heuristic scoring or GPT-4 scoring) to measure the feedback adherence of the trained model responses compared to the response of the base model for in-scope prompts. The in-scope adherence score for an algorithm on a given piece of feedback is the average of the per-prompt feedback scores ($h(x,y,\bar y)$ if the prompt $x$ is heuristically checkable, $g(x,y,\bar y)$ otherwise) across all in-scope test prompts for that feedback. The overall in-scope adherence score $S_\text{in}$ is the average of these per-feedback adherence scores over all feedbacks evaluated.
\begin{figure}
    \centering
    \includegraphics[width=\columnwidth]{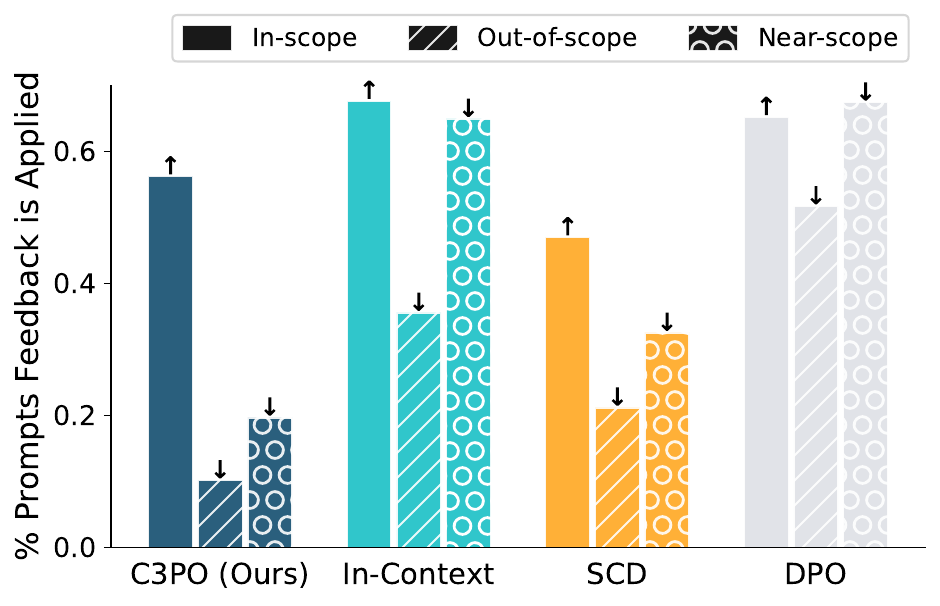}
    \caption{\textbf{{\method} substantially reduces overgeneralization} (applying the given feedback to prompts where it is not actually relevant) with only minor reduction in adherence to feedback for prompts where the feedback is relevant.}
    \label{fig:main}
\end{figure}

For out-of-scope prompts, our goal is to measure changes in the model's behavior as a result of incorporating the feedback. Because the model's behavior is most likely to change along the dimension of the feedback, we measure model behavior change on out-of-scope as the average \textit{change} in the rate we instead measure the absolute change in feedback adherence since over this domain, our objective is to leave the model behavior unchanged. To compute the out-of-scope behavior change score for an algorithm on a given piece of feedback, we average the \textit{absolute value} of the feedback scores (again, $h(x,y,\bar y)$ if $x$ is a heuristically checkable prompt, $g(x,y,\bar y)$ otherwise). That is, the average \textit{change} in feedback adherence over the set of out-of-scope test prompts. The overall behavior change score $S_\text{out}$ is the average of these per-feedback behavior change scores over all feedbacks evaluated. Apart from these individual metrics, we define $S_\text{overall} = \frac{S_\text{in} + \left(1 - S_\text{out}\right)}{2}$ as a combined metric with equal weighting of the in-scope feedback adherence score $S_\text{in}$ and one minus the out-of-scope behavior change score $S_\text{out}$. A $S_\text{overall}$ near one indicates that an algorithm effectively adheres to the given feedback better than the baseline model for in-scope prompts while preserving the level of feedback adherence in the baseline model on out-of-scope prompts.
\begin{table}
    \small
    \centering
    \resizebox{\columnwidth}{!}{\begin{tabular}{lcccc}
    \toprule
        \textbf{Method} & $S_\text{in}$ & $S_\text{out}$ & $S_\text{overall}$ \\
        \midrule
         In-Context & \textbf{0.677} $\pm$ \textbf{.028} & 0.503 $\pm$ .025 & 0.587 $\pm$ .026  \\
         In-Context + CoT & 0.402 $\pm$ .033 & 0.246 $\pm$ .017 & 0.578 $\pm$ .026  \\
         \midrule
         SCD & 0.470 $\pm$ .029 & 0.269 $\pm$ .019 & 0.6005 $\pm$ .025  \\
         SCD + Negatives & 0.367 $\pm$ .027 & \textbf{0.133} $\pm$ \textbf{.013} & 0.617 $\pm$ .021  \\
         \midrule
         DPO & 0.326 $\pm$ .048 & 0.517 $\pm$ .022 & 0.4045 $\pm$ .037  \\
         {\methodshort} (Ours) & 0.563 $\pm$ .031 & 0.150 $\pm$ .014 & \textbf{0.7065} $\pm$ \textbf{.024}  \\
         \bottomrule
    \end{tabular}}
    \caption{\textbf{{\methodshort} provides the strongest overall performance averaged over 100 pieces of feedback.} Augmentations to the In-Context and SCD baselines, In-Context + CoT and SCD + Negatives, respectively, reduce the amount of overgeneralization of both approaches, but at a substantial cost to in-scope feedback adherence, and ultimately these improvements do not change the overall score for either method.}
    \label{tab:main_results}
\end{table}

\textbf{Methods.} We compare {\methodshort} against both in-context-learning-based and fine-tuning methods. For \textbf{In-Context} learning, we provide the baseline model with the user query as well as the feedback and a prompt instructing the model to selectively apply the feedback whenever it is applicable to the given user query. We explore an enhancement of this naive prompting approach, \textbf{In-Context + CoT} using chain-of-thought prompting; this approach is the same as In-Context, except we prompt the model to first reason step-by-step about the applicability of the provided feedback to the given prompt before answering. See \cref{sec:appendix-prompts} for the full prompts. Next, we compare against performing supervised context distillation \textbf{SCD} on in-scope prompts, using the revised responses $y^+$ generated by {\methodshort} as the supervision targets. In addition, in order to better control overgeneralization, we evaluate \textbf{SCD + Negatives}, which adds a weighted constraint to the SCD loss over out-of-scope prompts, using the baseline model responses on these prompts as the supervision target. Finally, we evaluate \textbf{DPO} on the preference dataset $\mathcal{D}_\text{in-scope}$ without additional regularization, essentially an ablation of {\methodshort}'s regularization losses. For all experiments, we use Mistral-7B-Instruct-v0.2 \cite{jiang2023mistral} and train with Low-Rank Adaptation \cite{hu2021lora} with a rank of 64 and alpha of 128. We use a learning rate of 5e-5 with a cosine decay schedule and a warmup ratio of $0.05$ and train for 1 epoch.

\subsection{Quantifying and mitigating overgeneralization   }
Our initial results in \cref{fig:overgeneralization} show that for existing approaches to learning from verbal feedback, successfully incorporating the feedback (large x-axis value) leads to application of that feedback for prompts where the feedback does not apply (large y-axis value). That is, successful adherence to feedback for in-scope prompts comes at a high cost in terms of incorrect adherence to that feedback for out-of-scope prompts, shown by the large slope of the best fit lines. This result also shows that {\methodshort} trades off in-scope adherence and out-of-scope behavior preservation much more efficiently, shown by the much smaller slope of the best-fit line. In this section, we study the impact of incorporating verbal feedback in terms of the evaluation metrics $S_\text{in}$ and $S_\text{out}$, measuring out-of-scope behavior change for both generic out-of-scope prompts and more difficult near-scope prompts. The results are shown in \cref{fig:main}. {\methodshort} dramatically reduces the behavior change for both general out-of-scope prompts and near-scope prompts, while only slightly reducing feedback adherence for in-scope prompts.
\begin{figure}
    \centering
    \includegraphics[width=\columnwidth]{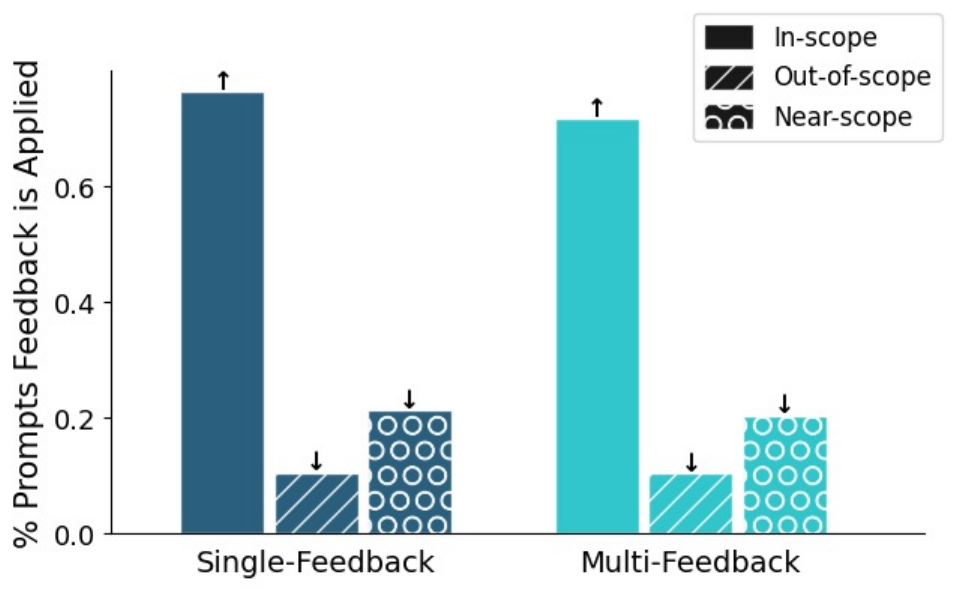}
    \caption{\textbf{Mixing the LoRA weights that {\methodshort} learns for two different pieces of feedback} (right) provides virtually the same level of in-scope feedback adherence and out-of-scope behavior change as applying and evaluating each feedback independently (left). This result suggests that learning separate LoRA parameters for each feedback as it is received and simply adding them together to acquire the model adapted to all feedbacks may be viable.}
    \label{fig:two_adapter_result}
\end{figure}

It is natural to wonder whether we can improve the performance of the baseline methods using a similar goal of constraining the behavior change for out-of-scope prompts. We therefore evaluate the modifications of the In-Context and SCD methods, In-Context + CoT and SCD + Negatives, intended to reduce behavior change. The In-Context + CoT method first performs chain-of-thought reasoning to decide whether the feedback is applicable to the given input before responding; the SCD + Negatives baseline mixes in the regularization loss on the $\mathcal{D}_\text{near-scope}$ and $\mathcal{D}_\text{out-of-scope}$ datasets. We report the in-scope feedback adherence and out-of-scope behavior change (averaged over the general out-of-scope and near-scope prompts) in \cref{tab:main_results}. While both improvements do substantially reduce the amount of overgeneralization compared to the original version of each method, they come at a substantial cost to in-scope feedback adherence. Therefore, the overall score $S_\text{overall}$ does not substantially improve for either of these approaches; however, {\methodshort} offers a more efficient tradeoff between in-scope feedback adherence and out-of-scope behavior change, shown by the significantly higher $S_\text{overall}$. We therefore conclude that {\methodshort} is an effective way to reduce overgeneralization while generally maintaining in-scope feedback adherence. This claim is supported by the results in \cref{fig:feedback_samples}.

\subsection{Adhering to multiple feedbacks}
So far, our experiments have evaluated the average result of applying a single piece of feedback to a pre-trained model. While a comprehensive evaluation of continual learning of many feedbacks is out of the scope of this work, we perform an initial investigation in this direction to assess the effectiveness of mixing the LoRA parameters learned from separate feedbacks. That is, we perform {\methodshort} separately on two different feedbacks $z_1$ and $z_2$, producing LoRA parameters $\phi_1$ and $\phi_2$. Ideally, to produce a model that adheres to \textit{both} $z_1$ and $z_2$, rather than re-training, we could simply use the mixture of residual parameters $\phi' = \phi_1 + \phi_2$. In this section, we compare the performance of this approach to combining feedback-adapted models. \cref{fig:two_adapter_result} compares the average feedback adherence and behavior change when applying and evaluating only a single piece of feedback at a time (left) with the average feedback adherence and behavior change on two feedbacks after applying the merged LoRA parameters $\phi'$ (right), averaged across 20 feedbacks (10 pairs). We observe virtually no degradation in in-scope feedback adherence and no change in out-of-scope behavior change. This result is promising for the possibility of enabling rapid, mix-and-match personalization and customization of large language models without re-training.

\subsection{Choice of {\methodshort} constraint formulation}

\begin{figure}
    \centering
    \includegraphics[width=\columnwidth]{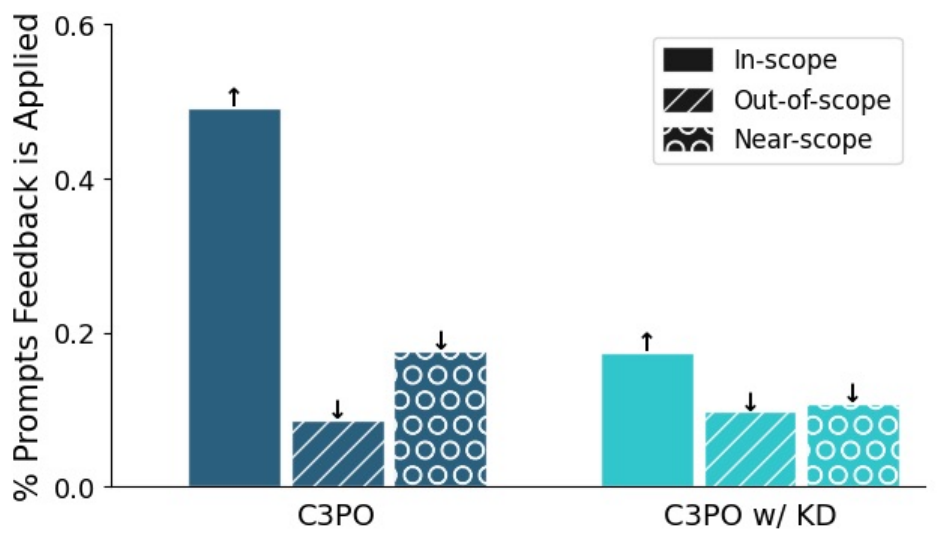}
    \caption{Replacing {\methodshort}'s maximum likelihood constraint on out-of-scope prompts with full knowledge distillation leads to substantially impaired performance on in-scope prompts, suggesting that allowing some subtle changes to the model's conditional distribution for out-of-scope prompts may be beneficial.}
    \label{fig:constraint_choice}
\end{figure}

{\methodshort} constrains the feedback-adapted model by maximizing the likelihood of the baseline model's responses on out-of-scope prompts during fine-tuning. We hypothesize that performing full knowledge distillation \citep{Hinton2015DistillingTK} for the conditional distributions of the adapted model and baseline model at each time step. That is, performing knowledge distillation on only out-of-scope prompts, where the baseline model is the teacher and the adapted model is the student. Rather than simply maximizing the likelihood of baseline model samples from each out-of-scope prompt, this constraint minimizes the KL-divergence between the baseline model's conditional distribution and the adapted model's conditional distribution, averaged over all timesteps in the baseline model's completion. While this form of constraint has successfully leveraged the `dark knowledge' represented by the lower-probability logits in the model's output to constrain neural networks in the context of continual learning \citep{buzzega2020dark}, we find that this stronger constraint substantially impairs in-scope feedback adherence compared to the maximum likelihood constraint. Alternative approaches to constraining out-of-scope model behavior is an important topic for future work.

\section{Discussion \& Future Work}
As large language models become more ubiquitous and widely-adopted, the ability to easily customize and personalize their behaviors is increasingly valuable. We therefore have studied the problem of learning from high-level verbal feedback, where a user provides a short piece of written feedback describing a desired change to a language model's behaviors, and an algorithm must update the model to adhere to this feedback when it is appropriate, but preserve the model's behaviors elsewhere. We showed that applying existing methods for fine-tuning language models from feedback demonstrate severe \textbf{overgeneralization}: after incorporating the feedback, they also change model behavior for inputs that are \textit{not} relevant to the provided feedback. To mitigate this problem, we introduced {\method} (\methodshort), an algorithm that performs reinforcement learning from verbal feedback (RLVF). {\methodshort} leverages existing language models to generate a set of small fine-tuning datasets that encode both the desired change in behavior described by the feedback and the set of behaviors that should be \textit{preserved} for inputs unrelated to the feedback. We found that {\methodshort} substantially reduces overgeneralization while still adhering to feedback for relevant inputs. Our experiments raise several important questions for future work. Can we perform continual learning from feedback by simply continually aggregating and mixing adapted model weights? In addition, our ablations of the {\methodshort} constraint loss function suggest that the proper level of `strictness' when constraining the model update is a non-trivial problem, and better-performing alternatives may exist. Finally, future work may investigate the relationship between the complexity of the feedback and the capabilities of the base model being adapted.

\section*{Acknowledgements}
The authors gratefully acknowledge the OpenAI Researcher Access Program and Modal.com for providing compute credits used in this project. CF is a CIFAR Fellow. EM acknowledges Kyle Hsu for helpful conversations.

\bibliography{icml2024/bibliography}
\bibliographystyle{icml2024}

\newpage
\appendix

\section{Sampling Details}
We sample from GPT-4 using a temperature of 0.7 and a top-p value of 0.7. When sampling from Mistral-7B-Instruct-v0.2, use a temperature of 0.7, a top-p value of 0.7, and top-k value of 50 and a repetition penalty of 1.

\section{Training Details}
To conduct our hyperparameter search, we select 10 arbitrary pieces of feedback from the human-generated feedback dataset. For all methods, we train using LoRA and choose a rank of 64, alpha of 128, and a LoRA dropout of 0.05. We observe that a smaller rank results in decreases in-scope feedback adherence while increasing the rank above 64 results in degradation of model completions across all methods.

We train for 1 epoch with a learning rate of $5\mathrm{e}{-5}$ using a cosine learning rate schedule and a $0.05$ warmup ratio. We ablate learning rates from $1\mathrm{e}{-7}$ to $1\mathrm{e}{-4}$ and found that below $1\mathrm{e}{-5}$, in-scope feedback adherence never increases sufficiently, even when training for multiple epochs. We, thus, picked our learning rate to be larger than this threshold but still small enough avoid a plateauing loss. In addition, we experimented with training for multiple epochs but found that this does not significantly alter the results and sometimes even increases feedback adherence for out-of-scope and near-scope prompts.

To choose $\lambda_1$ and $\lambda_2$ of the $\mathcal{L}_{\methodshort}$ objective, we conducted a grid search over the two hyperparameters. We found that it is crucial to set both $\lambda_1 > 0$ and $\lambda_2 > 0$ and that increasing $\lambda_2$, which is the weight for the SFT loss over the near-scope samples, beyond 0.1 decreases in-scope feedback adherence drastically while only marginally mitigating overgeneralization. Additionally, we found that \methodshort is not as sensitive over the choice of $\lambda_1$ as long as it is non-zero and chose $0.2$ due to the best in-scope to out-of-scope performance trade-off on our evaluation set.

For both \methodshort and DPO, we select a $\beta$ parameter of 0.1. We ablate with values of 0.05, 0.15, and 0.25 and find that a setting of 0.05 results in less in-scope feedback adherence and more overgeneralization while $\beta > 0.1$ only reduces in-scope feedback adherence.

\section{Derivation of Optimal Policy for PbRL on Two-Policy Preference Pairs}
\label{sec:math}
In this section, we derive several properties of the PbRL learning procedure used by {\methodshort}. First, we demonstrate that the synthetic preference data generated by {\methodshort} adheres to the commonly-used Bradley-Terry model \citep{bradley1952rankanalysis} of discrete choice and compute the true reward function implied by the preference data. Using this reward function, we then show that the optimal policy learned by {\methodshort} for in-scope prompts takes the form in \cref{eq:2p-policy}. Finally, we perform a simple empirical validation of this theoretical result in a synthetic bandit problem.

\subsection{Deriving the underlying Bradley-Terry scoring function for synthetic two-policy preference data}
The Bradley-Terry model of discrete choices states that for a preference over two responses $y$, $y'$, we have
\begin{equation}
    \label{eq:2p-bt}
    p(y \succ y' | x,y,y') = \sigma \left(r^*(x,y) - r^*(x,y')\right)
\end{equation}
for some true scoring function $r^*$. In the case of two-policy preference data, we assume that the preferred response is generated from a policy $\pi^+$ and the dispreferred response is generated by a policy $\pi^-$. The probability of observing the response pair $y,y'$ is $p(y,y'|x,A)=\pi^+(y|x)\pi^-(y'|x)$ and $p(y,y'|x,\neg A)=\pi^+(y'|x)\pi^-(y|x)$, where $A$ is the event that $y$ was generated by $\pi^+$ and $y'$ from $\pi^-$ (and therefore $y \succ y'$ by the definition of our data-generating process). By Bayes' Rule, the probability of $A$ (that $y$ was generated by $\pi^+$ and $y'$ by $\pi^-$) is
\begin{multline}
    \label{eq:2p-data-bayes}
    p(A|x,y,y') = p(y \succ y' | x,y,y') = \\\frac{\pi^+(y|x)\pi^-(y'|x)}{\pi^+(y|x)\pi^-(y'|x) + \pi^+(y'|x)\pi^-(y|x)}.
\end{multline}
We now set the RHS of \cref{eq:2p-bt} equal to the RHS of \cref{eq:2p-data-bayes} and show that a Bradley-Terry scoring function exists. We have:
\begin{multline}
    \sigma \left(r^*(x,y) - r^*(x,y')\right) = \\\frac{\pi^+(y|x)\pi^-(y'|x)}{\pi^+(y|x)\pi^-(y'|x) + \pi^+(y'|x)\pi^-(y|x)}.
\end{multline}
Applying the transformation $\log \frac{z}{1-z}$ to both sides, we have
\begin{equation}
    r^*(x,y) - r^*(x,y') = \log \frac{\pi^+(y|x)}{\pi^-(y|x)} - \log \frac{\pi^+(y'|x)}{\pi^-(y'|x)},
\end{equation}
implying the Bradley-Terry scoring function
\begin{equation}
    \label{eq:2p-scoring-function}
    r^*(x,\bar{y}) = \log \frac{\pi^+(\bar{y}|x)}{\pi^-(\bar{y}|x)} + C(x)
\end{equation}
for some value $C(x)$ that is constant with respect to $y$. Therefore, synthetic two-policy preference data is Bradley-Terry, with the above scoring function.

\subsection{Deriving the optimal policy for {\methodshort} for in-scope prompts}
The optimal policy for the KL-regularized reinforcement learning objective optimized by DPO is shown in prior work \citep{peters2007reinforcement,peng2019advantage,rafailov2023direct} to be
\begin{equation}
    \pi^*(y | x) = \frac{1}{Z(x)}\pi_\text{ref}(y | x)\exp\left(\frac{1}{\beta}r(x,y)\right).
\end{equation}
Substituting the optimal reward function in \cref{eq:2p-scoring-function}, assuming $C(x)=0$ as this constant term does not affect the optimal policy, we have
\begin{align}
    \pi^*(y | x) &= \frac{1}{Z(x)}\pi_\text{ref}(y | x)\exp\left(\frac{1}{\beta}\log \frac{\pi^+(y|x)}{\pi^-(y|x)}\right) \\
    &= \frac{1}{Z(x)}\pi_\text{ref}(y | x) \left(\frac{\pi^+(y|x)}{\pi^-(y|x)}\right)^{\frac{1}{\beta}}.
\end{align}
Assuming that $\pi_\text{ref}=\pi^-$ gives
\begin{align}
    \pi^*(y | x) &= \frac{1}{Z(x)}\pi^-(y | x) \frac{\pi^+(y|x)^{\frac{1}{\beta}}}{\pi^-(y|x)^{\frac{1}{\beta}}} \\
    &= \frac{1}{Z(x)}\frac{1}{\pi^-(y | x)^{-\frac{\beta}{\beta}}} \frac{\pi^+(y|x)^{\frac{1}{\beta}}}{\pi^-(y|x)^{\frac{1}{\beta}}} \\
    &\propto \left(\frac{\pi^+(y | x)\phantom{^{1-\beta}}}{\pi^-(y | x)^{1-\beta}}\right)^\frac{1}{\beta},
\end{align}
which is the optimal policy expression given in \cref{eq:2p-policy}.

\subsection{Empirically validating the in-scope {\methodshort} policy in a synthetic setting}

\begin{figure}
    \includegraphics[width=\columnwidth]{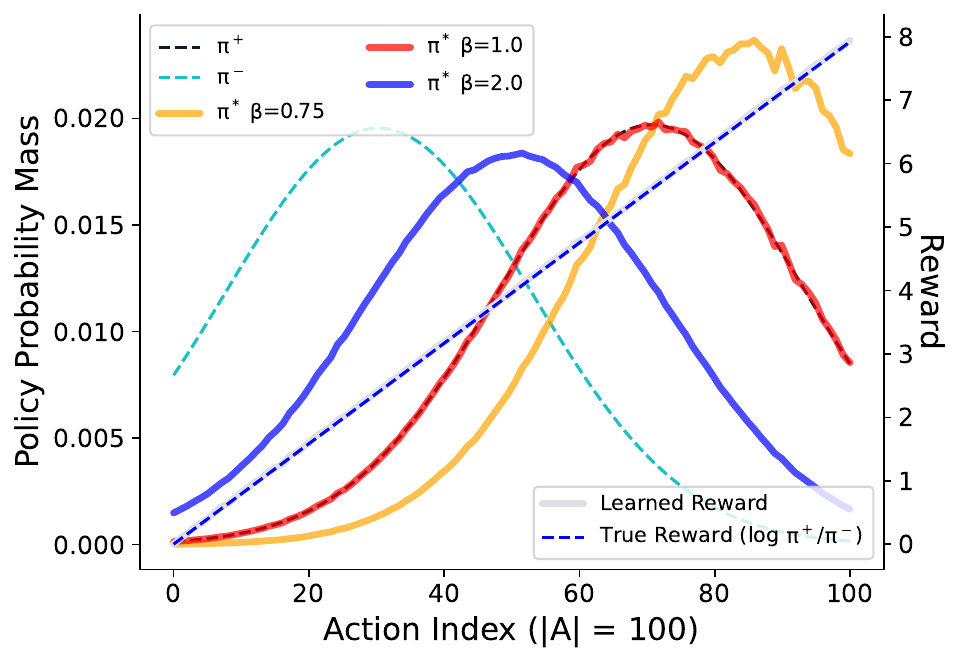}
    \label{fig:synthetic}
    \caption{Empirical demonstration of the result of {\methodshort}'s optimal policy on in-scope prompts. For $\beta=1$, the optimal policy recovers $\pi^+$; for smaller beta, the optimal policy amplifies the delta between $\pi^+$ and $\pi^-$. Note that the reward learned with the Bradley-Terry loss perfectly recovers the theoretically optimal reward in \cref{eq:2p-scoring-function}.}
\end{figure}
We validate the theoretical results in the previous subsections in a simple bandit setting with 100 possible actions. We define the preferred policy $\pi^+$ as $\pi^+(i) \propto \exp \left(\frac{(i-70)^2}{1000}\right)$ and similarly define $\pi^-$ as the shifted policy $\pi^-(i) \propto \exp \left(\frac{(i-40)^2}{1000}\right)$. We generate 1e8 preference pairs $(y^+, y^-)$, where $y^+\sim\pi^+(\cdot)$ and similarly for $y^-$. We fit a reward function to this preference data using the Bradley-Terry loss, using 400 steps of gradient descent with the Adam optimizer \citep{KingBa15} with learning rate 1.0, initializing the reward function to the zero vector.

The results are shown in \cref{fig:synthetic}. We find that the recovered reward function is exactly the Bradley-Terry scoring function predicted by \cref{eq:2p-scoring-function}, and that the special cases predicted by \cref{eq:2p-policy} hold (e.g., with $\beta=1$, we have simply $\pi^* = \pi^+$). With $\beta<1$, we have the intuitive behavior of amplifying the delta between $\pi^+$ and $\pi^-$.

\onecolumn

\section{Prompts}
\label{sec:appendix-prompts}
The following sections contain the verbatim prompts used for the various stages of the {\methodshort} synthetic data generation procedure.

\subsection{Category Generation Prompt}

\begin{lstlisting}
You are a helpful assistant. You are helping a user come up with categories around a topic that will be used to create some questions that the topic applies to and some questions that the topic does not apply to.

Given a topic, come up with {count} creative and diverse categories that are an only slightly larger superset of the topic. Be creative, think out of the box, and keep the categories closely related to the topic. Ensure that for each category, it would be possible to easily come up with questions that are not related to the provided topic. Do not repeat categories and make sure you cover all relevant categories.

You should first respond with "THOUGHTS" that describe what you should and should not mention in your categories and why. Then output "CATEGORIES: " on a new line and output each category on a new line as part of a numbered list. Finally, output "REVISED_CATEGORIES: " on a new line followed a revised version of each of the categories you came up with. Use the revision to modify a category if it would be very hard to come up with a prompt for that category that the topic does not apply to. The revision should also be a numbered list. If you do a great job, you will be tipped $200.

--EXAMPLE 1--
TOPIC: current fashion trends

THOUGHTS: I should list categories that are either related to fashion but that are not explicitly about trends. None of the categories I respond with should be directly about fashion trends.

CATEGORIES:
1. Buying luxury fashion
2. gen-z pop culture trends
3. fast-fashion trends
4. men's attire
5. planning an outfit
...

REVISED_CATEGORIES:
1. Buying luxury fashion
2. gen-z pop culture trends
3. fast-fashion
4. men's attire
5. planning an outfit
...
-- END EXAMPLE 1--

--EXAMPLE 2--
TOPIC: social media direct messages

THOUGHTS: I could list categories related to social media or messaging in general. Any category that includes but is broader than social media messaging is fine.

CATEGORIES:
1. Communicating with strangers via DM
2. Complimenting a friend
3. Sharing a post with a friend
4. Interacting with creators
5. Making plans to meet a friend
...

REVISED_CATEGORIES:
1. Communicating with strangers
2. Complimenting a friend
3. Sharing a post with a friend
4. Interacting with creators
5. Making plans to meet a friend
...
-- END EXAMPLE 2--

TOPIC: {topic}

THOUGHTS:
\end{lstlisting}

\subsection{In-Scope Sampling Prompt}
\begin{lstlisting}
You are a helpful assistant that always closely follows instructions. You are provided with a topic, and category. Your job is to come up with {count} actionable prompts that fulfill the following criteria:

- All prompts must fall within the category provided
- All prompts must be phrased in a way that both the prompt and eventual response will ALWAYS BE WITHIN the topic
- If a human had to modify all responses that fall within the topic, your prompts must be so clearly within the topic that the human would always have to make edits

Be very creative, think outside the box, and feel free to make up facts, names, and events to make the prompts more specific and actionable. Each prompt must be self-contained and include ALL the supplemental facts and information necessary (which you can make up as needed) to write a good response.

Each prompt should only be 1-3 sentences long. Do not repeat prompts and respond with NOTHING ELSE THAN THE PROMPTS. Output each prompt on a new line as part of a numbered list. If you do a great job, you will be tipped _200.

-- EXAMPLE 1--

TOPIC: the quality of airbus airplanes

CATEGORY: plane crashes

PROMPTS:
1. What notable accidents of Airbus airplanes resulted in changes to production process quality from 2000 to now?
2. Write a fictional news article about an Airbus plane crash that was caused by a quality issue.
3. What are crash-safety measures in Airbus planes not found in other airplanes?
4. Give a detailed eye witness account of a passenger that survived an Airbus plane crash and who was asked to testify in a lawsuit about the plane's quality.
5. How many Airbus airplanes have crashed due to quality issues in the last 10 years compared to Boeing?
6. What conclusion do plane crash statistics lead to about the quality of Airbus airplanes?
...
-- END EXAMPLE 1--

-- EXAMPLE 2--

TOPIC: texting my boss Jared

CATEGORY: asking for clarification on a task

PROMPTS:
1. Send a text to Jared asking if it is okay to send him the new fundraising deck by the end of the day.
2. Ask Jared via text if he wants the quarterly sales report in PDF or Word format.
3. Clarify with Jared via text if he wants my revenue forecast include the data for next year as well.
4. Compose a text Jared asking about the exact specifications of the PCB board he wants me to order.
...
-- END EXAMPLE 2--

TOPIC: {domain}

CATEGORY: {category}

PROMPTS:
\end{lstlisting}

\subsection{Near-Scope Sampling Prompt}

\begin{lstlisting}
You are a helpful assistant that always closely follows instructions. You are provided with a topic to avoid and a category. Your job is to come up with {count} example prompts that fulfill the following criteria:

- All prompts must fall within the category provided
- All prompts must not fall within the provided topic to avoid but closely related (if there is some intersection between the category and topic, focus your prompts on the aspects of the category that is not part of the topic)
- If a human had to modify all responses that fall within the topic to avoid, your prompts must be so clearly outside the topic that the human would never have to make any edits

Be EXTREMELY creative, think outside the box, and MAKE UP ANY facts, names, and events to make the prompts more specific, actionable, and realistic. Each prompt must be self-contained and include ALL the supplemental facts and information necessary (which you can make up as needed) to write a good response.

Each prompt should only be 1-3 sentences long. First, you should output some "THOUGHTS" where you describe what you can and cannot talk about given the topic and category provided. Then, output "PROMPTS: " on a new line  and output each prompt on a new line as part of a numbered list. Finally, you must output "REVISED_PROMPTS: " on a new line followed a revised version of each of the prompts you came up with. Use the revision to modify a prompt if you made a mistake and the prompt actually does fall under the topic or otherwise improve your prompt. The revision should also be a numbered list. If you do a great job, you will be tipped _200.

--EXAMPLE--
TOPIC_TO_AVOID: the quality of airbus airplanes

CATEGORY: plane crashes

THOUGHTS: I need to come up with prompts related to plane crashes but I am not allowed to talk about the quality of Airbus airplanes. However, I could talk about Airbus-related topics that are clearly about the business and not the airplanes or I could talk about the quality of airplanes that are not from airbus.

PROMPTS:
1. What are notable accidents of Boeing airplanes from 2000 to now?
2. Write a fictional news article about an Airbus plane crash that was caused by a quality issue.
3. What business segments of Airbus operate in the satellite industry?
4. What air plane manufacturers are there apart from Boeing and Airbus?
5. Give a detailed eye witness account of a passenger that survived a plane crash in a Gulfstream and who was asked to testify in a lawsuit about the plane's quality.
6. What is the safety record of Embraer airplanes vs. Airbus?
7. What is the chance of survival in a plane crash?
8. You are the CEO of Boeing. Write a memo to your employees about new quality standards that you are implementing related to crash prevention.
9. Write insurance ad copy for a company that insures Boeing airplanes.
...

REVISED_PROMPTS:
1. What are notable accidents of Boeing airplanes from 2000 to now?
2. Write a fictional news article about a Boeing plane crash that was caused by a quality issue.
3. What business segments of Airbus operate in the satellite industry?
4. What air plane manufacturers are there apart from Boeing and Airbus?
5. Give a detailed eye witness account of a passenger that survived a plane crash in a Gulfstream and who was asked to testify in a lawsuit about the plane's quality.
6. What is the safety record of Embraer airplanes?
7. What is the chance of survival in a plane crash?
8. You are the CEO of Boeing. Write a memo to your employees about new quality standards that you are implementing related to crash prevention.
9. Write insurance ad copy for a company that insures Boeing airplanes.
...
-- END EXAMPLE--

TOPIC_TO_AVOID: {domain}

CATEGORY: {category}

PROMPTS:
\end{lstlisting}

\subsection{In-Context + CoT Prompt}

\begin{lstlisting}
You are a helpful assistant. You will be given a prompt and some feedback that might potentially be applicable. 
Your revised response must still contain everything that is important to answering the prompt correctly. 
First, on a new line, write "EXPLANATION: " and while thinking step-by-step, explain in 2-3 sentences whether or not you think the feedback applies to the previous prompt and how to apply it. 
Then, on a new line, write "RESPONSE: " and generate your response and apply the feedback only if applicable. 
Do not output anything besides the response after your response.

PROMPT: {prompt}
FEEDBACK: {feedback}
EXPLANATION: 
\end{lstlisting}

\subsection{Revise Completion Prompt}

\begin{lstlisting}
You are a helpful assistant. You are given a prompt, a previous response, and some feedback. Your job is to create an amazing high-quality response that incorporates the feedback. Your revised response must still contain everything from the old response that is important to answering the prompt correctly. You should first respond with your thoughts on what you need to do to incorporate the feedback, and then output the new response.

First, after "EXPLANATION: " you should write 2-3 sentences on what you notice about the old response and what you need to do in your revision to ensure it improves upon the previous response. Make sure to think step-by-step, so your revision is as good as possible. Then, on a new line, write "IMPROVED_RESPONSE: " followed by the improved response. DO NOT OUTPUT ANYTHING ELSE AFTER THE IMPROVED RESPONSE.

PROMPT: {prompt}

PREVIOUS_RESPONSE: {response}

FEEDBACK: {feedback}

EXPLANATION: 
\end{lstlisting}

\subsection{Evaluate Feedback Adherence Prompt}
\begin{lstlisting}
You are a helpful assistant. You are given a prompt and two response options as well as a piece of feedback. Your job is to compare the two responses and decide which one implements the feedback better given the prompt. Your response should be on a scale from 1 to 5 where each score has the following meaning:

1: RESPONSE_1 implements the feedback much better than RESPONSE_2
2: RESPONSE_1 implements the feedback better than RESPONSE_2
3: Both responses implement the feedback equally well
4: RESPONSE_2 implements the feedback better than RESPONSE_1
5: RESPONSE_2 implements the feedback much better RESPONSE_1

First, after "EXPLANATION: " you should write 2-3 sentences on what you notice about the two responses and why one might implement the feedback better than the other. Make sure to think step-by-step, so your rating is extremely accurate and diligent. 
Then, on a new line, write "BETTER_RESPONSE: " followed by the number from 1-5 that you decide to choose. DO NOT OUTPUT ANYTHING ELSE AFTER THE NUMBER.

PROMPT: {prompt}

RESPONSE_1: {completion1}

RESPONSE_2: {completion2}

FEEDBACK: {feedback}

EXPLANATION:
\end{lstlisting}

\subsection{Evaluate Completion Helpfulness Prompt}

\begin{lstlisting}
You are a helpful assistant. You are given a prompt and two response options. Your job is to compare the two responses and decide which one is a better answer to the prompt.

Your response should be on a scale from 1 to 5 where each score has the following meaning:

1: RESPONSE_1 is much better than RESPONSE_2
2: RESPONSE_1 is better than RESPONSE_2
3: Both responses answer the prompt equally well
4: RESPONSE_2 is better than RESPONSE_1
5: RESPONSE_2 is much better RESPONSE_1

First, after "EXPLANATION: " you should write 2-3 sentences on what criteria you think a good prompt should fulfill, what you notice about the two responses, and why one might be better than the other. Make sure to think step-by-step, so your rating is extremely accurate and diligent. Then, on a new line, write "BETTER_RESPONSE: " followed by the score from 1-5 that you decide to choose. DO NOT OUTPUT ANYTHING ELSE AFTER THE NUMBER.


PROMPT: {prompt}

RESPONSE_1:
"{completion1}"

RESPONSE_2:
"{completion2}"

EXPLANATION:
\end{lstlisting}

\subsection{Feedback Generation Prompt – Style}
\begin{lstlisting}
You are a helpful assistant that always closely follows instructions. Your overall task is to generate feedback which a user gives to a LLM to improve its responses. The user is asking the LLM for responses and has found something they would like the LLM to improve upon. This means the feedback should be something a LLM would not already follow well. For example, feedback to "write work emails more politely" is NOT GOOD because LLMs already generate very polite work emails. The feedback should target something the LLM can improve on. Assume this LLM only takes text as input and outputs only text.

Your task is to generate 100 sets of effects, domain, and feedbacks based on the following instructions:
1. Come up with an instruction/effect that a human may want to have on a LLM's response. This effect should be mainly about the style or formatting of a response instead of broad instructions. The effect should not focus on content or tone. 
2. Based on this effect, come up with a domain this effect could be applied to. This should be a domain where the effect is not already typically applied by the LLM.
3. Combine the effect and domain to create a piece of feedback. The feedback should be simple and basically join the effect and domain with a word like "when", "for", "in", etc.

Below are a few examples:
Example 1: { "effect"="use London gangster vernacular", "domain": "sending texts to my friend Peter", "feedback": "Use London gangster vernacular when texting my friend Peter" }
Example 2: { "effect"="be more detailed", "domain": "writing an email to my PI Anna", "feedback": "Be more detailed in your emails to my PI Anna" }
Example 3: { "effect"="be more concise", "domain": "writing an email to my boss Jared", "feedback": "Be more concise when emailing my boss Jared" }
Example 4: { "effect"="end emails with "Best,\nMoritz"", "domain": "writing work emails", "feedback": "End work emails with "Best,\nMoritz"" }
Example 5: { "effect"="use German", "domain": "writing emails to my colleague Max", "feedback": "Use German when emailing my colleague Max" }

Be creative and think out of the box. Do not repeat feedback, effects, or domains. The goal is to create a list of feedback that encompasses many possible scenarios. Output ONLY the feedback, effect, and domain in structured json format.
\end{lstlisting}

\subsection{Feedback Generation Prompt – Content}
\begin{lstlisting}
You are a helpful assistant that always closely follows instructions. Your overall task is to generate feedback which a user gives to a LLM to improve its responses. The user is asking the LLM for responses and has found something they would like the LLM to improve upon. This means the feedback should be something a LLM would not already follow well. For example, feedback to "write work emails more politely" is NOT GOOD because LLMs already generate very polite work emails. The feedback should target something the LLM can improve on. Assume this LLM only takes text as input and outputs only text.

Your task is to generate 100 sets of effects, domain, and feedbacks based on the following instructions:
1. Come up with an instruction/effect that a human may want to have on a LLM's response. This effect be mainly about the content or tone of a response. The effect should not focus on style or formatting. 
2. Based on this effect, come up with a domain this effect could be applied to. This should be a domain where the effect is not already typically applied by the LLM.
3. Combine the effect and domain to create a piece of feedback. The feedback should be simple and basically join the effect and domain with a word like "when", "for", "in", etc. Do not provide extra information beyond the effect and domain.

Below are a few examples:
Example 1: { "effect": "assume that I do not have any knowledge of math beyond a fourth grade level", "domain": "Explaining math questions" , "feedback": "When explaining math questions, assume that I do not have any knowledge of math beyond a fourth grade level"} 
Example 2: { "effect": "make stories have dramatic, negative endings", "domain": "Doing creative writing", "feedback": "Whenever you do creative writing, make sure that stories have dramatic, negative endings" } 
Example 3: { "effect": "use simple language and sentences", "domain": "writing children's books", "feedback": "When writing children's books, use simple language and sentences" } 
Example 4: { "effect": "use a refined and controlled tone", "domain": "texting my mom", "feedback": "When texting my mom, use a refined and controlled tone"} 
Example 5: { "effect": "use a formal and professional tone", "domain": "emailing Jared", "feedback": "When emailing Jared, use a formal and professional tone"}

Be creative and think out of the box. Do not repeat feedback, effects, or domains. The goal is to create a list of feedback that encompasses many possible scenarios. Output ONLY the feedback, effect, and domain in structured json format.
\end{lstlisting}

\section{Feedback examples}
\cref{tab:example-prompts} shows examples of feedback from humans and GPT-4. Adherence to `qualitative' feedback must be measured with a language model like GPT-4, while adherence to `quantitative' feedback can be checked with a simple heuristic or regular expression.

\begin{table}[H]
    \small
    \centering
    \begin{tabular}{p{1.5cm}p{5.5cm}}
        \toprule
        Human Qualitative & Always use some heart or kiss emoji when texting my girlfriend Maddie \\
        \midrule
        \multirow{2}{*}{\shortstack[l]{Human\\Quantitative}} & Use the term `lol' when responding to text messages from friends \\
        \midrule
        GPT-4 & Use alliteration creatively for brand naming suggestions \\
        \bottomrule
    \end{tabular}
    \caption{Examples of verbal feedback from the human and GPT-4-generated datasets in our experiments.}
    \label{tab:example-prompts}
\end{table}

\section{Sample Results}

\textbf{Feedback}: Utilize a F.A.Q. format in writing informational brochures for a healthcare clinic
\\
\textbf{In-Context Prompt}: Write a section for a brochure discussing the healthcare clinic's partnership with local schools for health education programs.
\\
\textbf{Baseline}
\begin{lstlisting}
**Empowering the Next Generation: Our Partnership with Local Schools**

At [Healthcare Clinic Name], we believe that a healthy community starts with a healthy foundation. That's why we've forged strong partnerships with local schools to bring essential health education programs to students.

Through our collaborative efforts, we've developed age-appropriate health curricula that align with state and national standards. Our team of experienced healthcare professionals and educators work together to deliver engaging and informative sessions on topics such as:

* Proper nutrition and healthy eating habits
* Importance of regular exercise and physical activity
* Mental health awareness and stress management
* Personal hygiene and disease prevention
* Substance abuse prevention
* First aid and emergency preparedness

These interactive programs are designed to foster a lifelong commitment to wellness among students. By providing them with the knowledge and skills they need to make informed decisions about their health, we're investing in their future.

Our partnership with local schools also extends beyond the classroom. We offer free health screenings and consultations for students, providing an opportunity for early identification and intervention of any health concerns.

Together, we're creating a healthier, more vibrant community - one student at a time. Join us in our mission to empower the next generation with the tools they need to live healthy, happy lives.
\end{lstlisting}

\textbf{In-Context}
\begin{lstlisting}
**Q:** **How does our healthcare clinic collaborate with local schools for health education programs?**

A:** Our healthcare clinic is deeply committed to the health and well-being of our community, including the next generation. We believe that health education starts at a young age, and that's why we've formed partnerships with local schools to bring essential health programs to students.

**Q:** **What types of health education programs does the clinic offer in schools?**

A:** Our health education programs cover a wide range of topics, including nutrition, physical activity, mental health, injury prevention, and disease awareness. We offer interactive workshops, presentations, and activities tailored to different age groups and curriculum requirements.

**Q:** **Who leads these health education programs in schools?**

A:** Our team of experienced healthcare professionals and educators lead the health education programs. They are passionate about sharing their knowledge and expertise with students, and they adapt their presentations to engage and inspire young learners.

**Q:** **How often do these health education programs take place in schools?**

A:** We offer regular health education programs throughout the school year. Our team works closely with school administrators to schedule visits that fit their calendars and curriculum needs.

**Q:** **Is there a cost for these health education programs in schools?**

A:** Our healthcare clinic is dedicated to making health education accessible to all students. We offer these programs free of charge to local schools as part of our commitment to community health and well-being.

**Q:** **How can schools get involved with the healthcare clinic's health education programs?**

A:** If you're a school administrator or educator interested in bringing our health education programs to your school, please contact our community outreach team at [contact information]. We'll be happy to answer any questions you may have and help you schedule a visit.

**Q:** **What other ways does the healthcare clinic support local schools and students?**

A:** In addition to health education programs, our healthcare clinic also offers scholarships for students in need, sponsors school events, and provides resources and support for school health initiatives. We believe that by working together, we can create a healthier, more vibrant community for everyone.
\end{lstlisting}

\textbf{SCD}
\begin{lstlisting}
**Q: What is the partnership between [Clinic Name] and local schools about?**
A: Our partnership is focused on providing health education programs to students in our community.

**Q: Why is health education important for students?**
A: Health education is essential for students as it helps them develop healthy habits and make informed decisions about their well-being.

**Q: What health education programs does [Clinic Name] offer in schools?**
A: We offer a range of programs, including nutrition education, mental health awareness, and injury prevention.

**Q: How does [Clinic Name] engage students in these programs?**
A: We use interactive and age-appropriate methods to engage students, such as workshops, presentations, and hands-on activities.

**Q: What benefits do students gain from these programs?**
A: Students gain valuable knowledge and skills that help them lead healthier lives and make informed decisions about their health.

**Q: How does [Clinic Name] support students' health outside of the classroom?**
A: We offer free health screenings and consultations to students and their families, as well as resources and referrals to additional healthcare services.

**Q: How can parents and guardians get involved in these programs?**
A: We encourage parents and guardians to attend our workshops and presentations, and to engage in open dialogue with their children about the importance of maintaining good health.

**Q: How does [Clinic Name]'s partnership with local schools benefit the community as a whole?**
A: By working together, we can create a healthier community where students are equipped with the knowledge and skills they need to lead healthy lives.

**Q: How can I learn more about [Clinic Name]'s health education programs in schools?**
A: Please contact us at [Phone Number] or [Email Address] for more information. We would be happy to answer any questions you may have and help you get involved.
\end{lstlisting}

\textbf{DPO}
\begin{lstlisting}
**Q: What is the goal of the health education programs?**
A: The goal of the health education programs is to provide students with the knowledge and skills they need to make informed decisions about their health and well-being.

**Q: What topics are covered in the health education programs?**
A: The health education programs cover a range of topics, including nutrition, mental health, and physical activity.

**Q: How often do the health education programs take place?**
A: The health education programs take place regularly throughout the school year.

**Q: Who leads the health education programs?**
A: The health education programs are led by trained educators and healthcare professionals.

**Q: How can I get involved in the health education programs?**
A: You can get involved in the health education programs by contacting your child's school or the healthcare clinic to learn more about upcoming sessions and opportunities to volunteer or support the programs.

**Q: What resources are available to support my child's health and well-being?**
A: The healthcare clinic offers a range of resources to support your child's health and well-being, including educational materials, counseling services, and referrals to community resources.

**Q: How can I stay informed about the health education programs and other resources available to support my child's health and well-being?**
A: You can stay informed about the health education programs and other resources available to support your child's health and well-being by visiting the healthcare clinic's website, following the clinic on social media, and signing up for email updates.

**Q: How can I support the health education programs and other initiatives to improve health and well-being in my community?**
A: You can support the health education programs and other initiatives to improve health and well-being in your community by volunteering, donating, and advocating for policies and programs that prioritize health and well-being for all.

**Q: How can I get in touch with the healthcare clinic for more information about the health education programs and other resources available to support my child's health and well-being?**
A: You can get in touch with the healthcare clinic by calling (555) 555-5555 or visiting the clinic's website to learn more about the health education programs and other resources available to support your child's health and well-being.
\end{lstlisting}

\textbf{\methodshort}
\begin{lstlisting}
**Q: What health education programs does [Clinic Name] offer in partnership with local schools?**
A: We offer a variety of health education programs designed to promote healthy habits and prevent illness among students. These programs include:

**Q: How does [Clinic Name]'s health education programs benefit students?**
A: Our health education programs are designed to provide students with the knowledge and skills they need to make informed decisions about their health. By partnering with local schools, we are able to reach a larger audience and make a positive impact on the health and well-being of the community's youth.

**Q: What topics are covered in [Clinic Name]'s health education programs?**
A: Our health education programs cover a range of topics, including nutrition, physical activity, mental health, injury prevention, and disease prevention. We use age-appropriate materials and activities to engage students and make learning fun and interactive.

**Q: How can I get involved with [Clinic Name]'s health education programs?**
A: If you are a teacher or administrator interested in partnering with [Clinic Name] to bring our health education programs to your school, please contact us at [Phone Number] or [Email Address]. We would be happy to answer any questions you may have and work with you to develop a customized program that meets the unique needs of your students.

**Q: What is the goal of [Clinic Name]'s health education programs?**
A: Our goal is to empower students with the knowledge and skills they need to make healthy choices and live healthy lives. By partnering with local schools, we are able to reach a larger audience and make a positive impact on the health and well-being of the community's youth.

**Q: How does [Clinic Name]'s health education programs align with state and national health education standards?**
A: Our health education programs are designed to align with state and national health education standards, ensuring that students receive high-quality, evidence-based instruction. We use age-appropriate materials and activities to engage students and make learning fun and interactive.

**Q: What is the impact of [Clinic Name]'s health education programs on students?**
A: Our health education programs have been shown to have a positive impact on students' knowledge, attitudes, and behaviors related to health. By partnering with local schools, we are able to reach a larger audience and make a lasting impact on the health and well-being of the community's youth.

**Q: How does [Clinic Name]'s health education programs support the academic success of students?**
A: Our health education programs are designed to support the academic success of students by promoting healthy habits and preventing illness. By
\end{lstlisting}

\textbf{Near-Context Prompt}: Discuss the development of an AI chatbot that helps patients navigate their health insurance options.

\textbf{Baseline}
\begin{lstlisting}
Developing an AI chatbot to help patients navigate their health insurance options is an innovative solution to address the complexity and confusion often associated with health insurance plans. Here's a step-by-step discussion on how to develop such a chatbot:

1. Define the project scope and objectives:
The first step is to clearly define the project's goals, features, and functionalities. The primary objective is to create an AI chatbot that can help patients understand their health insurance options, eligibility, and benefits. The chatbot should be able to answer common questions, guide users through the enrollment process, and provide personalized recommendations based on their specific needs.

2. Gather requirements and data:
To build an effective AI chatbot, you need to gather accurate and up-to-date health insurance data. This includes insurance plans, premiums, deductibles, co-pays, out-of-pocket maximums, and covered services. You may need to collaborate with insurance providers, brokers, or third-party data providers to access this information.

3. Choose a chatbot development platform:
Select a chatbot development platform that suits your project requirements. Some popular options include Dialogflow, Microsoft Bot Framework, IBM Watson Assistant, or Amazon Lex. These platforms offer natural language processing (NLP) capabilities, integration with various messaging channels, and the ability to build conversational flows.

4. Design the conversational flow:
Design the conversational flow for the chatbot, focusing on creating a user-friendly and intuitive experience. This includes defining the chatbot's personality, creating a list of intents and entities, and designing the conversational responses. Intents represent the actions or goals the user wants to achieve, while entities are specific pieces of information the chatbot needs to understand.

5. Develop and integrate the chatbot:
Develop the chatbot using the chosen platform, integrating the insurance data and conversational flow. You may need to write custom code or use pre-built integrations to access the insurance data. Ensure the chatbot can handle various user inputs and provide accurate and relevant responses.

6. Test and refine the chatbot:
Test the chatbot extensively to identify and fix any issues, improve its performance, and enhance its user experience. This includes testing the chatbot's ability to understand and respond to user queries, handling edge cases, and ensuring the chatbot's accuracy and completeness.

7. Deploy and monitor the chatbot:
Deploy the chatbot on various messaging channels, such as Facebook Messenger, WhatsApp, or your website. Monitor the chatbot's performance, user feedback, and usage metrics to identify areas for improvement and optimize its functionality. Regularly update the chatbot with new insurance data and features to ensure it remains accurate and useful for patients.

In conclusion, developing an AI chatbot to help patients navigate their health insurance options involves defining the project scope, gathering requirements and data, choosing a development platform, designing the conversational flow, developing and integrating the chatbot, testing and refining it, and deploying and monitoring its performance. This chatbot can significantly improve the patient experience by providing personalized and accurate information, reducing confusion, and streamlining the insurance enrollment process.
\end{lstlisting}

\textbf{In-Context}
\begin{lstlisting}
Title: AI Chatbot "HealthNav": Your Personal Health Insurance Navigator

1. **Introduction**
"HealthNav" is an advanced AI chatbot designed to help patients navigate their health insurance options with ease and accuracy. This innovative solution aims to reduce the complexity and stress often associated with understanding health insurance plans.

2. **Features**
- Instant responses to common health insurance queries
- Personalized recommendations based on user's insurance plan and medical history
- Integration with various insurance providers for real-time information
- User-friendly interface for seamless interaction

3. **Development Process**
**Step 1:** Defining the Scope
The first step was to identify the key features and functionalities required for the AI chatbot. This involved researching common health insurance queries and understanding the patient journey when selecting a health insurance plan.

**Step 2:** Designing the Conversational Flow
The conversational flow was designed using a combination of rule-based and machine learning algorithms. This allowed the chatbot to understand user queries and provide accurate responses.

* Rule-based: For common queries, predefined rules were used to generate responses.
* Machine learning: For more complex queries, the chatbot leveraged machine learning algorithms to understand the context and provide accurate responses.

**Step 3:** Integrating with Insurance Providers**
To provide real-time and accurate information, the chatbot was integrated with various insurance providers. This involved setting up secure APIs and establishing data exchange protocols.

**Step 4:** Testing and Iteration**
Extensive testing was carried out to ensure the chatbot provided accurate and helpful responses. Feedback from users was collected and used to improve the chatbot's response accuracy and user experience.

4. **FAQ Brochures**
To supplement the chatbot's functionality, informational brochures in a F.A.Q. format were created for patients. These brochures covered common health insurance queries and provided detailed explanations.

* Topics: Coverage details, deductibles, copays, premiums, and out-of-pocket maximums
* Format: Easy-to-understand language and visuals
* Accessibility: Available online and in print at the healthcare clinic

5. **Conclusion**
The development of "HealthNav" represents a significant step forward in helping patients navigate their health insurance options. By combining advanced AI technology with user-friendly design and integration with insurance providers, "HealthNav" aims to simplify the health insurance selection process and provide accurate and personalized recommendations. Additionally, the F.A.Q. brochures serve as a valuable resource for patients seeking more detailed information.
\end{lstlisting}

\textbf{SCD}
\begin{lstlisting}
Title: Navigating Health Insurance: An FAQ on Developing an AI Chatbot for Patients

Q: What is an AI chatbot, and how can it help patients navigate their health insurance options?
A: An AI chatbot is a computer program designed to simulate human conversation. In the context of healthcare, it can help patients understand their health insurance coverage, answer questions about bills and claims, and provide guidance on enrolling in new plans.

Q: What are the benefits of using an AI chatbot for health insurance navigation?
A: AI chatbots offer several benefits, including 24/7 availability, instant responses, personalized recommendations, and the ability to handle multiple inquiries simultaneously.

Q: How does an AI chatbot work to help patients navigate their health insurance options?
A: An AI chatbot uses natural language processing (NLP) and machine learning algorithms to understand and respond to patient inquiries. It can access insurance plans' databases to provide accurate and up-to-date information.

Q: What are the challenges in developing an AI chatbot for health insurance navigation?
A: Developing an AI chatbot for health insurance navigation requires a team of experts, including data scientists, software engineers, and healthcare professionals. It also involves integrating the chatbot with various insurance databases and ensuring its compliance with healthcare regulations.

Q: How can an AI chatbot improve patient engagement and satisfaction?
A: An AI chatbot can improve patient engagement and satisfaction by providing personalized recommendations, answering questions in real-time, and reducing the need for lengthy phone calls or in-person visits.

Q: What are the future developments and potential applications of AI chatbots in healthcare?
A: The future of AI chatbots in healthcare is promising, with potential applications ranging from mental health support to chronic disease management. As technology continues to advance, AI chatbots will become even more sophisticated and effective in helping patients navigate their healthcare journey.
\end{lstlisting}

\textbf{DPO}
\begin{lstlisting}
**Q: What is the purpose of the AI chatbot?**
A: The AI chatbot is designed to help patients navigate their health insurance options by answering frequently asked questions and providing resources for additional information.

**Q: What is the chatbot's primary function?**
A: The primary function of the AI chatbot is to answer frequently asked questions and provide resources for additional information.

**Q: How does the chatbot work?**
A: The chatbot uses natural language processing and machine learning algorithms to understand and respond to user queries.

**Q: What is the chatbot's availability?**
A: The chatbot is available 24/7 to answer user queries.

**Q: How does the chatbot provide resources for additional information?**
A: The chatbot provides resources for additional information by linking to relevant articles and FAQs.

**Q: How does the chatbot ensure accuracy and up-to-date information?**
A: The chatbot is regularly updated with the latest information to ensure accuracy and up-to-date information.

**Q: How does the chatbot handle complex queries?**
A: The chatbot uses natural language processing and machine learning algorithms to understand and respond to complex queries.

**Q: How does the chatbot handle errors and misunderstandings?**
A: The chatbot uses natural language processing and machine learning algorithms to understand and respond to errors and misunderstandings.

**Q: How does the chatbot handle user feedback and suggestions?**
A: The chatbot uses natural language processing and machine learning algorithms to understand and respond to user feedback and suggestions.

**Q: How does the chatbot handle user privacy and security?**
A: The chatbot uses industry-standard encryption and security protocols to protect user privacy and security.

**Q: How does the chatbot handle user authentication and authorization?**
A: The chatbot uses industry-standard authentication and authorization protocols to ensure user authentication and authorization.

**Q: How does the chatbot handle user accessibility and usability?**
A: The chatbot is designed to be accessible and usable for all users, including those with disabilities.

**Q: How does the chatbot handle user experience and design?**
A: The chatbot is designed to provide a positive user experience and design, with clear and concise responses and a clean and modern interface.

**Q: How does the chatbot handle user engagement and retention?**
A: The chatbot is designed to engage and retain users by providing personalized and relevant responses and a clear and concise interface.

**Q: How does the chatbot handle user satisfaction and feedback?**
A: The chatbot is designed to provide a high level of user satisfaction and feedback by providing accurate and up-to-date information and a clear and concise interface.

**Q: How does the chatbot handle user support and troubleshooting?**
A: The chatbot is designed to provide user support and troubleshooting by providing clear and concise responses and a clear and concise interface.

**Q: How does the chatbot handle user integration and customization?**
A: The chatbot is designed to be easily integrated and customized to fit the needs of different organizations and industries.

**Q: How does the chatbot handle user scalability and performance?**
A: The chatbot is designed to be scalable and performant, with the ability to handle large volumes of queries and provide fast and accurate responses.

**Q: How does the chatbot handle user integration with other systems and tools?**
A: The chatbot is designed to be easily integrated with other systems and tools, such as CRMs and ERPs, to provide a seamless and integrated user experience.

**Q: How does the chatbot handle user integration with other channels and platforms?**
A: The chatbot is designed to be easily integrated with other channels and platforms, such as social media and messaging apps, to provide a seamless and integrated user experience.

**Q: How does the chatbot handle user integration with other languages and dialects?**
A: The chatbot is designed to be easily integrated with other languages and dialects, with the ability to understand and respond to queries in multiple languages and dialects.

**Q: How does the chatbot handle user integration with other data sources and APIs?**
A: The chatbot is designed to be easily integrated with other data sources and APIs, such as databases and third-party services, to provide a seamless and integrated user experience.

**Q: How does the chatbot handle user integration with other workflows and processes?**
\end{lstlisting}

\textbf{\methodshort}
\begin{lstlisting}
Developing an AI chatbot to help patients navigate their health insurance options is an innovative solution to address the complexity and confusion often associated with health insurance plans. Here's a discussion on the key steps involved in developing such a chatbot:

1. Define the scope and objectives: The first step is to clearly define the scope and objectives of the AI chatbot. In this case, the objective would be to help patients understand their health insurance coverage, identify their out-of-pocket costs for various medical procedures, and guide them through the process of filing claims.

2. Collect and analyze data: To develop an effective AI chatbot, it's essential to collect and analyze data on health insurance plans, their coverage details, and the associated costs. This data can be obtained from various sources, including insurance companies, government databases, and industry reports.

3. Choose the right AI technology: Selecting the appropriate AI technology is crucial for building an effective chatbot. Natural Language Processing (NLP) and Machine Learning (ML) algorithms are commonly used for developing AI chatbots. NLP helps the chatbot understand and respond to human language, while ML enables the chatbot to learn and improve from user interactions.

4. Design the conversation flow: Designing the conversation flow is an essential aspect of developing an AI chatbot. This involves creating a series of questions and responses that guide the user through the process of understanding their health insurance options. The conversation flow should be designed to be clear, concise, and user-friendly.

5. Develop the chatbot: Once the data has been collected, the AI technology chosen, and the conversation flow designed, the chatbot can be developed. This involves writing the code, integrating the AI algorithms, and building the user interface.

6. Test and refine the chatbot: After the chatbot has been developed, it's essential to test it thoroughly and refine it based on user feedback. This involves identifying and addressing any bugs or issues, improving the conversation flow, and enhancing the user experience.

7. Deploy and monitor the chatbot: Once the chatbot has been tested and refined, it can be deployed on various platforms, such as websites, mobile apps, or messaging platforms. It's essential to monitor the chatbot's performance regularly and make any necessary updates or improvements based on user feedback and data analysis.

In conclusion, developing an AI chatbot to help patients navigate their health insurance options involves defining the scope and objectives, collecting and analyzing data, choosing the right AI technology, designing the conversation flow, developing the chatbot, testing and refining it, and deploying and monitoring it. This solution can significantly improve the patient experience by making it easier for them to understand their health insurance coverage and manage their medical expenses.
\end{lstlisting}

\section{Full List of Feedbacks Used}
\subsection{Manually Generated}
\begin{itemize}
    \item Always use some heart or kiss emoji when texting my girlfriend Maddie
    \item Use '\&' instead of 'and' in any Slack message DMs to my colleagues John, Michael, Eric, or Hailey
    \item Be more concise when emailing my boss Jared
    \item For specific Python coding questions (about syntax, popular library use etc.), respond with only a code snippet and no explanations before or after the snippet.
    \item Use a more casual tone in work emails to people on my team
    \item When writing a Haiku, always use rhymes
    \item Explaining anything related to quantum physics or relativity as if you were talking to a 9-year-old.
    \item Assume that your audience is PhD students and use highly technical language when writing about concepts related to artificial intelligence
    \item When talking about HIV/AIDS in Rwanda, make sure the first sentence has a 1st word of 'The'
    \item Use sports analogies when writing motivational emails to the sales team
    \item Whenever you do creative writing ensure that your stories have dramatic, negative, grim endings.
    \item When writing messages to my parents, include some German phrases
    \item When asked for advice on how to deal with difficult life situations, always include a lighthearted but appropriate joke
    \item Do not use greetings in text messages to my friends
    \item Be very casual in work Slack messages
    \item Include equations when explaining concepts related to machine learning
    \item Always assert that Techno is the best music genre when writing about music
    \item Do not use buzzwords or technical jargon when writing about startups
    \item When asked a computer science question, offer only a very brief high level overview and ask the user what part of the answer they want to learn more about.
    \item When answering questions that require a code snippet but the desired language is not mentioned, always write the code snippet in Elixir.
    \item When asked about advice on fashion choices, give an extremely firm, one-sided answer
    \item For anything related to dinosaurs, only answer in gibberish
    \item When talking about cars, speak like a pirate.
    \item For any questions related to calculus, do not respond with an answer but instead say that the user should already know the answer and do not say anything else.
    \item Whenever you talk about Biology, use one or two relevant emojis
    \item Be more concise when writing comments on social media
    \item When writing science fiction stories, to make them dinosaur themed
    \item Do not use any heart or kiss emojis when commenting on male friends' social media posts
    \item When helping me resolve conflicts with people, always use 'I feel' statements
    \item Do not use any hashtags when commenting on Instagram posts
    \item When writing song lyrics, do not use any commas, periods, exclamation marks, or question marks
    \item When responding to social media posts, sound like a cringey LinkedIn influencer.
    \item Finish emails to my boss Sam with 'Cheerio, Sasha'
    \item When you send invites for meetings with more than 3 participants, keep the meeting length to at most 30 minutes
    \item Use the term 'lol' when responding to text messages from friends
    \item When helping me come up with a workout plan for Mondays and Tuesdays, limit the number of unique exercises to three
    \item When talking about my parents, always call them by their first names, 'Doris' and 'Michael'.
    \item When giving recommendations on vacation destinations, always suggest places in Germany
    \item When helping me schedule meetings on Tuesdays and Thursdays, make sure I have at least a 15 minute break in between activities
    \item When asked about a potential menu for restaurants based in San Francisco, only suggest Asian fusion dishes
    \item When scheduling meetings that are not with my boss, never schedule them before 11am
    \item When talking about historic events, always format your answer in three parts 'Background:', 'Event:', and 'Consequences:'
    \item When asked for advice on good finance books, include the \emojibooks \emojibooks emoji in your response.
    \item When giving recommendations for wedding venues in California, only mention places directly on the ocean
    \item Assume that I am talking about a Golden Retriever when I ask you for pet health advice
    \item When writing cold outreach emails for networking, make a lot of puns
    \item When sending work emails about deliverables to your direct reports over the weekend, do not give them any new TODO's
    \item When recommending recipes for family dinners, only give me vegetarian recipes
    \item When answering questions related to weather, remember that I live in Alaska
    \item When giving me fitness advice, use the voice of a military drill sergeant
\end{itemize}

\subsection{GPT4-Generated}
\begin{itemize}
    \item When reporting on financial news, frame information with historical context
    \item When reviewing technology products, discuss potential ethical dilemmas
    \item When discussing conspiracy theories, adopt a skeptical tone
    \item When explaining legal jargon, create relatable analogies
    \item When describing consumer products, emphasize long-term impacts
    \item In addressing social issues, propose unconventional solutions
    \item When designing a university syllabus, invoke a casual, conversational tone
    \item When explaining local traditions, include a global perspective
    \item When writing news articles, embrace speculative fiction elements
    \item When analyzing music, draw connections to other art forms
    \item Format responses as haiku when creating social media captions
    \item Include ASCII art in customer support emails
    \item Insert relevant emojis when replying to online food reviews
    \item Format answers as multiple choice questions when responding to customer queries
    \item Write as if speaking to a child when composing messages for pediatric healthcare
    \item Utilize a F.A.Q. format in writing informational brochures for a healthcare clinic
    \item Write using rhythmic patterns when crafting social media posts for a music festival
    \item Use cooking metaphors when writing a newsletter for a culinary school
    \item Synthesize multiple languages in one response when replying to a cosmopolitan audience on social media
    \item Simulate an interview format in writing articles featuring professionals
    \item Draw parallels to historical events when explaining current news topics
    \item Portray a futuristic AI persona when composing tweets for a tech product launch
    \item Use a Socratic method of asking questions when explaining philosophy topics on a discussion forum
    \item Compose using iambic pentameter when writing custom greetings for greeting cards
    \item Insert hyperlinks to sources in blog posts on health topics
    \item Use alliteration creatively for brand naming suggestions
    \item Write in second person for a personal touch in customer service emails
    \item Apply a telegram style for brevity in emergency notifications
    \item When writing educational texts on finance, create a glossary of terms
    \item Highlight key points with bold text in executive summaries
    \item Apply satire and irony in political commentary
    \item Format as a diary entry for first-person narrative pieces
    \item Write with a mock-serious tone in pet product reviews
    \item Provide analogies and metaphors in self-help articles
    \item Implement bulleted lists when listing product features
    \item Write in hexameter for epic fantasy verses
    \item Use vernacular English in country music lyrics
    \item When writing travel advice, focus deeply on cultural sensitivities
    \item When teaching chemistry concepts, offer analogies with cooking
    \item When tutoring in physics, anticipate common misconceptions
    \item When crafting public service announcements, demonize negative behaviors subtly
    \item When teaching social studies, present facts in a storytelling format
    \item When giving motivational speeches, mimic the style of classical orators
    \item When explaining abstract math theorems, suggest practical, real-world applications
    \item When describing business case studies, highlight decision-making processes
    \item When recapping movie plots, suggest alternative ending scenarios
    \item In teaching history lessons, draw parallels to pop culture
    \item In crafting marketing copy, incorporate elements of suspense
    \item When writing speeches, use humor to illustrate points
    \item When covering medical breakthroughs, emphasize the ethical dilemmas
\end{itemize}

\end{document}